\documentclass[11pt, letterpaper]{article}

\usepackage[left=1in, right=1in, top=1in, bottom=1in]{geometry}
\usepackage[utf8]{inputenc}
\usepackage[T1]{fontenc}
\usepackage{bm}
\usepackage{type1cm}
\usepackage{lettrine}
\usepackage{amsmath,amssymb,amsthm}
\usepackage{moreverb}
\usepackage{mathtools}
\usepackage{amsmath}
\usepackage{amssymb}
\usepackage{algorithmic}
\usepackage{graphics}
\usepackage{graphicx}
\usepackage{subfigure}
\usepackage{caption}
\usepackage{extarrows}
\usepackage{color}
\usepackage{framed}
\usepackage{wrapfig}
\usepackage{bm}
\usepackage{mathrsfs}
\usepackage{mathabx}
\usepackage{multirow}
\usepackage{longtable}
\usepackage{hyperref}
\usepackage{paralist}
\usepackage{indentfirst}
\usepackage{relsize}
\usepackage{extarrows}
\usepackage{upgreek}
\usepackage{bm}
\usepackage{mwe}
\usepackage[dvipsnames,table]{xcolor}
\usepackage{booktabs}
\usepackage{authblk}
\usepackage{lettrine}
\usepackage{type1cm}
\usepackage{threeparttable}
\usepackage[sort&compress,numbers]{natbib}
\usepackage[figurename=Figure]{caption}
\usepackage[normalem]{ulem}

\graphicspath{ {./Figure/} }
\usepackage[font=footnotesize,labelfont=bf]{caption}

\providecommand{\keywords}[1]{\textbf{\textit{Keywords: }} #1}

\hypersetup{
bookmarks=true,
bookmarksopen=true,
bookmarksnumbered=true,
unicode=false,
pdftoolbar=true,
pdfmenubar=true,
pdffitwindow=false,
pdfstartview={FitH},
pdftitle={My title},
pdfauthor={Author},
pdfsubject={Subject},
pdfcreator={Creator},
pdfproducer={Producer},
pdfkeywords={keywords},
pdfnewwindow=true,
colorlinks=true,
linkcolor=blue,
citecolor=blue,
filecolor=blue,
urlcolor=blue
}

\usepackage{tikz}
\newcommand{\StartMenu}{\includegraphics[height=.018\linewidth]{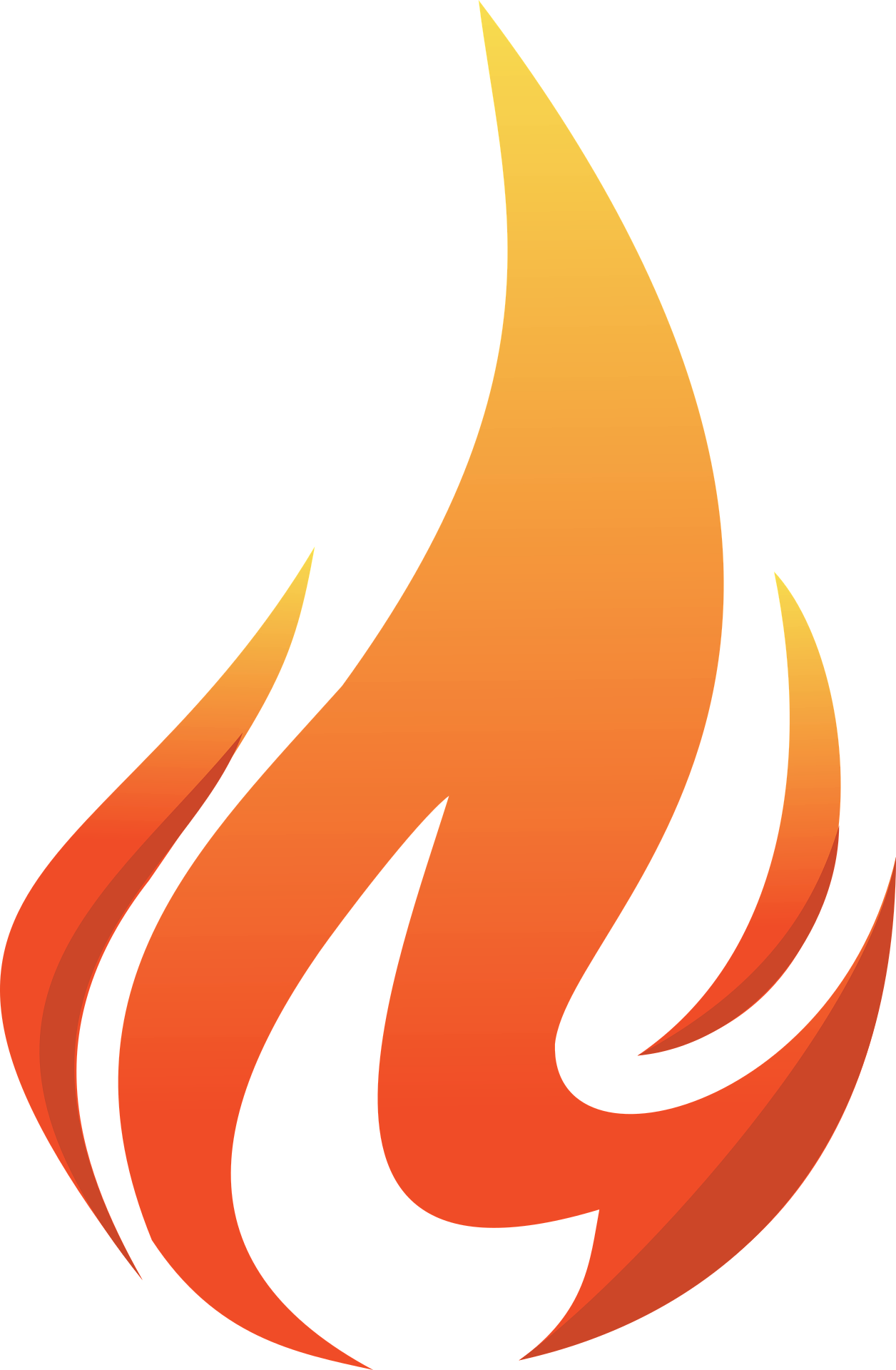}}%
\newcommand{\WinMenu}[1]{%
   \StartMenu
}%

\begin{document}

\title{\textbf{Protecting your portraits: Real-time Identity Defense against malicious personalization of diffusion models}}

\author[1,2]{Hanzhong Guo}
\author[1,2]{Shen Nie}
\author[3]{Chao Du}
\author[3]{Tianyu Pang}
\author[1,2$^*$]{Hao Sun}
\author[1,2$^*$]{Chongxuan Li}

\affil[1]{\small Gaoling School of Artificial Intelligence, Renmin University of China, Beijing, China}
\affil[2]{\small Beijing Key Laboratory of Big Data Management and Analysis Methods, Beijing, China} 
\affil[3]{\small Sea AI Lab, Singapore \vspace{18pt}}
\affil[*]{Correspondence to Hao Sun and Chongxuan Li}

\date{}

\maketitle

\normalsize

\vspace{-18pt} 
\begin{abstract}
	\small 
Personalized generative diffusion models, capable of synthesizing highly realistic images based on a few reference portraits, may pose substantial social, ethical, and legal risks via identity replication. Existing defense mechanisms rely on computationally intensive adversarial perturbations tailored to individual images, rendering them impractical for real-world deployment. This study introduces the Real-time Identity Defender (RID), a neural network designed to generate adversarial perturbations through a single forward pass, bypassing the need for image-specific optimization. RID achieves unprecedented efficiency, with defense times as low as 0.12 seconds on a single NVIDIA A100 80G GPU (4,400 times faster than leading methods) and 1.1 seconds per image on a standard Intel i9 CPU, making it suitable for edge devices such as smartphones. Despite its efficiency, RID achieves promising protection performance across visual and quantitative benchmarks, effectively mitigating identity replication risks. Our analysis reveals that RID’s perturbations mimic the efficacy of traditional defenses while exhibiting properties distinct from natural noise, such as Gaussian perturbations. To enhance robustness, we extend RID into an ensemble framework that integrates multiple pre-trained text-to-image diffusion models, ensuring resilience against black-box attacks and post-processing techniques, including image compression and purification. Our model is envisioned to play a crucial role in safeguarding portrait rights, thereby preventing illegal and unethical uses.
\end{abstract}


\vspace{12pt} 

\section*{Introduction} 

\begin{figure*}[t!]
\begin{center}
\centerline{\includegraphics[width=0.99\linewidth]{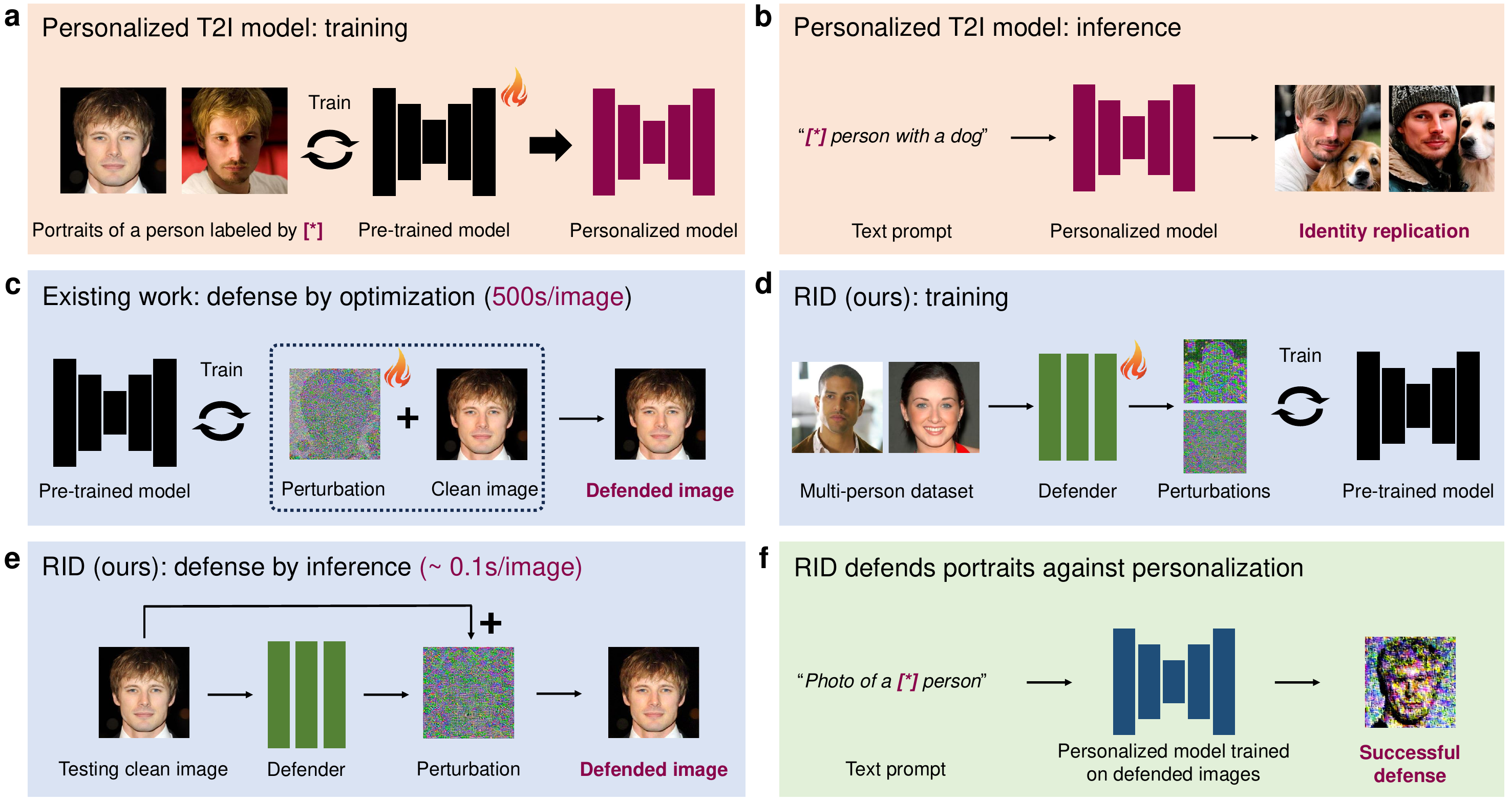}}
\caption{\textbf{Overview of personalized T2I diffusion models and our real-time identity defender (RID).} In all panels, the flame icon \WinMenu~~indicates trainable parameters. \textbf{a}, Personalized T2I models learn personal identity efficiently by fine-tuning a pre-trained TI2 model (e.g., Stable Diffusion~\cite{sd15,sd21,podellsdxl}) on a few portraits. \textbf{b}, Personalized T2I models can generate high-fidelity images by combining the learned identity and other concepts following the text prompt. \textbf{c}, Existing defense methods optimize an individual adversarial perturbation for each image against Personalized T2I techniques. \textbf{d}, RID trains a defender on a face dataset of multiple persons via adversarial score distillation sampling and regularization (detailed in the Method section). \textbf{e}, RID defends a new testing image by generating the corresponding perturbation through an efficient forward pass. \textbf{f}, RID successfully prevents personalized T2I techniques from learning personal identities in terms of visual perception. All facial images used in this figure are sourced from publicly available datasets and are permitted for academic purposes.}
\label{fig-sf}
\end{center}
\vspace{-6mm}
\end{figure*}

Large-scale diffusion models have made significant strides in generating high-fidelity images from textual descriptions, driven by their ability to generalize in a zero-shot manner—synthesizing novel combinations of concepts learned from extensive pretraining datasets~\cite{ramesh2021zero,rombach2022high,saharia2022photorealistic,betker2023improving,podellsdxl,esser2024scaling} (see Fig.~\ref{fig-sf}\textbf{a}).
Building on this foundation, \emph{personalization} techniques~\cite{kumari2023multi,dong2022dreamartist,han2023svdiff,xiang2023closer,xiao2023fastcomposer,liu2023facechain} inspired by \emph{delta tuning}~\cite{ding2022delta} have enabled diffusion models to replicate unique concepts, including individual identities, from as few as a dozen of reference images while maintaining generalization capabilities (see Fig.~\ref{fig-sf}\textbf{b}). This functionality has unlocked immense creative potential and found widespread applications in design, education, and entertainment.

However, the ability to replicate individual identities raises profound ethical, legal, and societal concerns. By fine-tuning open-source models like Stable Diffusion~\cite{sd15,sd21,podellsdxl}, malicious actors can generate highly realistic but harmful content, including disinformation or fabricated media, threatening personal privacy and public trust. The accessibility of pre-trained diffusion models and the efficiency of personalization techniques amplify these risks, enabling identity replication achievable within 15 minutes using consumer-grade hardware and publicly available tools. Given the vast number of images uploaded online daily, the potential for unauthorized identity replication and manipulation is vast and urgent.
  
To mitigate these threats, existing defense mechanisms~\cite{shan2023glaze,liang2023adversarial,madry2017towards,van2023anti,wang2023simac,salman2023raising} optimize protective adversarial perturbations~\cite{goodfellow2015explaining,madry2017towards}  for individual images, preventing diffusion models from learning the depicted identities\footnote{This is a kind of adversarial attack for good, i.e., leveraging the vulnerabilities of T2I models to protect identities.}. These perturbations, imperceptible to the human eye, are crafted to exploit vulnerabilities in diffusion models’ objective functions while adhering to norm constraints (see Fig.~\ref{fig-sf}\textbf{c}). Despite effective, these methods are computationally intensive and impractical for large-scale or real-time applications, as they require over 500 seconds per image to perform individual optimization.

To address these limitations, we propose the \emph{Real-time Identity Defender (RID)}, a neural network that generates adversarial perturbations through a single forward pass, eliminating the need for image-specific optimization. RID achieves unprecedented efficiency, processing images in as little as 0.12 seconds on a single NVIDIA A100 80G GPU (4,400 times faster than existing approaches) and 1.1 seconds per image on a standard Intel core i9 CPU, having potential to enable deployment on edge devices such as smartphones. Our approach is powered by a novel loss function, namely, \emph{Adversarial Score Distillation Sampling (Adv-SDS)}, which targets pre-trained diffusion models efficiently, and a regularization term that enhances the visual quality of defended images. RID is trained on a curated dataset of 70,000 multi-person images (see Fig.~\ref{fig-sf}\textbf{d}) and achieves promising protection performance in both visual perception and quantitative metrics while significantly improving efficiency and scalability. Our further analyses reveal that RID’s perturbations emulate the effectiveness of traditional defenses while remaining distinct from natural noise, such as Gaussian perturbations. 
 
Additionally, we extend RID with an ensemble framework that leverages multiple pre-trained diffusion models, ensuring robust defenses against black-box attacks and resilience to post-processing methods such as JPEG compression and diffusion-based purification. As a result, RID offers a scalable, practical solution to safeguarding identities in the era of generative personalization, bridging the critical gap between security and usability.

\section*{Results}

\subsection*{The Real-time Identity Defender (RID) Framework}

\paragraph{Pipeline.}  

RID introduces a neural network capable of generating protective perturbations for images in a single forward pass, eliminating the need for computationally intensive, image-specific optimization. As illustrated in Extended Data Fig.~\ref{fig-append2}, RID takes a clean image as input and outputs an adversarial noise. This noise is then added to the clean image to produce a protected image, which is subsequently fed into a pre-trained diffusion model that attempts to personalize it. Intuitively, the loss function of RID is designed to increase the difficulty of the personalization process, thereby achieving effective identity protection. After being trained on multi-person image datasets, RID will possess strong generalization capabilities, generating protective noise for new, unseen data through a single forward pass.

\paragraph{Training.}  

We propose adversarial score distillation sampling (Adv-SDS) loss as the primary training objective of RID, which optimizes the network to generate perturbations that disrupt the ability of personalized diffusion models to replicate protected identities. By adversarially maximizing the diffusion loss~\cite{ho2020denoising}, Adv-SDS ensures that defended images resist personalization while maintaining visual fidelity. This approach leverages the intrinsic properties of pre-trained diffusion models to craft effective perturbations, essentially achieving a robust trade-off between identity protection and image quality.

Technically, Adv-SDS addresses a significant computational challenge: directly calculating the Jacobian of large diffusion models with respect to their inputs is infeasible due to their complexity.  Drawing inspiration from score distillation sampling (SDS), which minimizes diffusion loss for zero-shot 3D generation~\cite{poole2022dreamfusion} rather than maximizing it for protection, we propose an efficient and theoretically grounded surrogate objective for Adv-SDS. This surrogate simplifies gradient computations while maintaining the defense’s effectiveness.

Although Adv-SDS produces promising identity protection, it occasionally introduces subtle grid-like artifacts into the defended images. To address this, we incorporate a regression loss that aligns RID’s perturbations with precomputed, image-specific perturbations from Anti-DB~\cite{van2023anti}. This regularization term enhances visual consistency and reduces artifacts without significantly increasing training complexity. However, our ablation study (See Extended Data Fig.~\ref{fig7}) reveals that the regression loss alone does not protect the identity effectively in isolation, highlighting its important role as a supplementary component.

Noteworthy, RID is trained on a curated dataset of 70,000 high-resolution images from the VGG-Face2 dataset~\cite{simonyan2014very}, selected following the established filtering criteria~\cite{chen2023simswap++}. The dataset includes diverse facial features, poses, and lighting conditions, ensuring robust generalization across varied scenarios.

\begin{figure*}[t]
\begin{center}
\centerline{\includegraphics[width=0.99\linewidth]{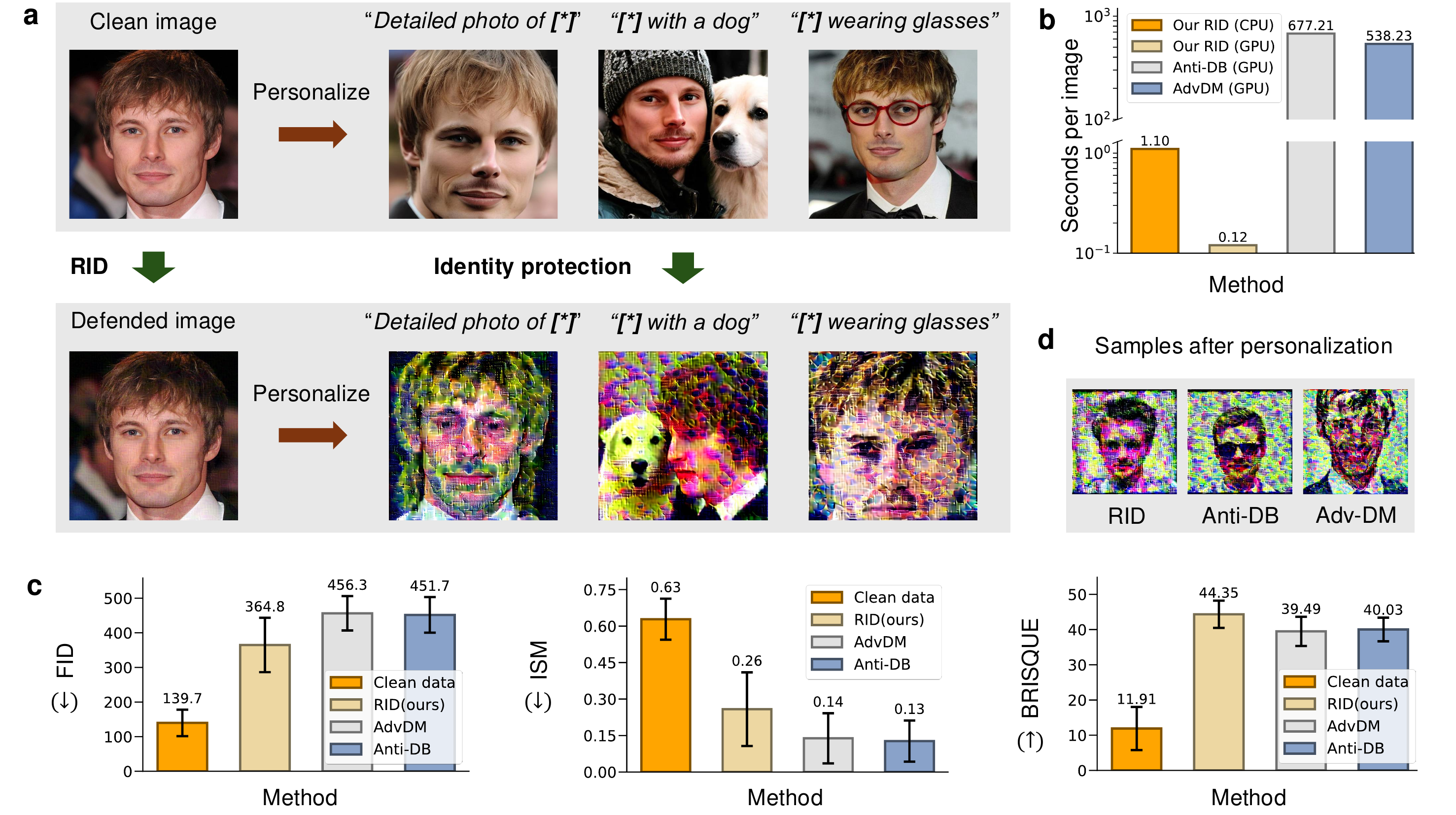}}
\caption{\textbf{RID defends the identities against malicious image
generation efficiently and effectively under various metrics.} \textbf{a}, The RID-defended image closely resembles the clean image (left column). Personalized diffusion models trained on clean images accurately retain identity across three prompts (top row), while models trained on RID-defended images produce distorted outputs with reduced identifiable features (bottom row).
\textbf{b}, RID achieves significantly faster defense speeds, with processing times of 1.1 seconds on an Intel I9 CPU and 0.12 seconds on a NVIDIA A100 80G GPU, compared to optimization-based methods such as Anti-DB and AdvDM, which require over 500 seconds on the same GPU.
\textbf{c}, Quantitative evaluation across three metrics (FID, ISM, and BRISQUE) shows that RID provides comparable protection to optimization-based methods and performs significantly better than the baseline without any defense. Arrows next to each metric denote better defense performance (i.e., lower visual quality).
\textbf{d}, Although the visual patterns of RID-defended samples differ from those of existing methods, RID achieves comparable defense performance qualitatively, consistent with the quantitative results depicted in \textbf{c}. All facial images used in this figure are sourced from publicly available datasets and are permitted for academic purposes.}
\label{fig2}
\end{center}
\vspace{-6mm}
\end{figure*}

\paragraph{Inference.} Once trained, RID generates protective perturbations for input images in a single forward pass. Each perturbation is directly added to the corresponding input image to create a defended version. This process is highly efficient, as shown in Fig.~\ref{fig2}\textbf{b}, achieving over three orders of magnitude speedup compared with existing methods (e.g., Anti-DB and AdvDM) under the same computational environment.

\subsection*{Evaluation Protocal}

\paragraph{Dataset and prompts.} For the evaluation, we used the open-source CelebA-HQ dataset~\cite{karras2017progressive}, different from the training set, to assess the generalization capabilities of RID. A total of 15 individuals (five females and ten males) were randomly selected, with 12 images per individual. During training, personalized diffusion models were fine-tuned on clean or RID-defended images using the prompt ``\textit{Photo of a [*] person}''. This consistent training prompt simulates real-world adversarial scenarios, where publicly available images are used to train models for identity replication. 

For inference, we generated outputs under two conditions: the same prompt used during training to assess alignment with training conditions for quantitative evaluation, and diverse prompts such as ``\textit{[*] person with a dog}'' to evaluate RID’s robustness in generalizing over novel scenarios. This two-fold testing approach ensures a comprehensive evaluation of RID’s defense performance across varying conditions.

\paragraph{Personalization methods.} To assess RID’s defense capabilities, we employed several personalization techniques. For the loss function, we adopted Dreambooth~\cite{ruiz2023dreambooth}, which preserves the generative generalization of personalized models. Two representative fine-tuning approaches were considered: the first involved training all parameters of the diffusion model (referred to as DB)~\cite{ruiz2023dreambooth}, while the second utilized a combination of two commonly used lightweight methods—textual embedding (TI)~\cite{gal2022image} and LoRA~\cite{hu2021lora} (denoted as LoRA+TI). DB was chosen for its strong personalization capability, whereas LoRA+TI serves as a widely adopted lightweight alternative. Notably, LoRA+TI typically outperforms TI or LoRA alone and is much more efficient than DB, offering an excellent balance between efficiency and effectiveness. Therefore, it is used as the default method throughout our experiments.

\paragraph{Quantitative metrics.} The quantitative evaluation focused on two key aspects, namely, identity protection and visual fidelity, using three common metrics. Frechet Inception Distance (FID)~\cite{heusel2017gans} measured the similarity between the distributions of generated and reference images, with higher FID scores for RID-defended images indicating a stronger defense. Identity Score Matching (ISM)~\cite{van2023anti} evaluated identity similarity using ArcFace~\cite{deng2019arcface} embeddings, where lower ISM scores reflect a better identity protection. Finally, the Blind/Referenceless Image Spatial Quality Evaluator (BRISQUE)~\cite{mittal2012no} assessed the perceptual image quality, with higher scores indicating a greater visual degradation. Details of the metrics are provided in the Method section.

\subsection*{Defending against malicious personalization}

To evaluate the effectiveness of RID, we conducted comprehensive qualitative and quantitative analyses, with the results shown in Figs.~\ref{fig2} and \ref{fig3}. These evaluations demonstrate RID’s ability to safeguard identities against personalization by text-to-image diffusion models while maintaining computational efficiency and adaptability across various settings.

\paragraph{Visual characteristics of defense results.}
Fig.~\ref{fig2}\textbf{a} illustrates the visual appearance of RID-defended images and their effectiveness in disrupting malicious personalization. The defended images retain a natural and realistic appearance, closely resembling the original clean inputs. When personalized diffusion models are fine-tuned on clean images, they produce realistic outputs that successfully preserve the subject’s identity across various prompts. In contrast, when fine-tuned on RID-defended images, the outputs become distorted and fail to reproduce identifiable facial features. This qualitative evidence highlights RID’s ability to disrupt identity replication without compromising the plausibility of the defended images.

\paragraph{Computational efficiency.}
Fig. \ref{fig2}\textbf{b}  compares the processing speeds of RID with representative optimization-based methods, such as Anti-DB~\cite{van2023anti} and AdvDM~\cite{liang2023adversarial}. While existing methods require over 500 seconds per image on a GPU, RID completes the defense process in just 0.12 seconds on a GPU and 1.1 seconds on a CPU. This makes RID approximately 4,400 times faster, possessing potential to enable practical deployment in real-time applications, such as on edge devices like smartphones, while maintaining a comparable defense performance.

\paragraph{Quantitative defense performance.}
Fig.~\ref{fig2}\textbf{c} presents a quantitative evaluation of RID’s defense effectiveness under three widely used metrics: FID, ISM, and BRISQUE, which collectively assess visual quality and defense robustness. RID achieves competitive performance across all three metrics, yielding comparable protection to optimization-based methods while significantly outperforming the baseline without defense. Although the the metrics in Fig.~\ref{fig2}\textbf{c} may show slightly weaker performance of RID compared with Anti-DB and AdvDM, a comparable level of protection is achieved, as demonstrated in Fig.~\ref{fig2}\textbf{d}, where visual inspection finds it difficult to distinguish the original image.
 
\paragraph{Comparison of visual patterns.} Fig.~\ref{fig2}\textbf{d} examines the visual patterns of RID-defended images in comparison to those protected by Anti-DB and AdvDM. Although the perturbation patterns generated by RID differ from those of optimization-based methods, all approaches achieve qualitatively effective protection, as evidenced by the distorted outputs shown in Fig.~\ref{fig2}\textbf{a}. This highlights the practical efficacy of RID’s defenses, even with its generalized, image-agnostic design.

\subsection*{Robustness of RID}

\begin{figure*}[t]
\begin{center}
\centerline{\includegraphics[width=0.99\linewidth]{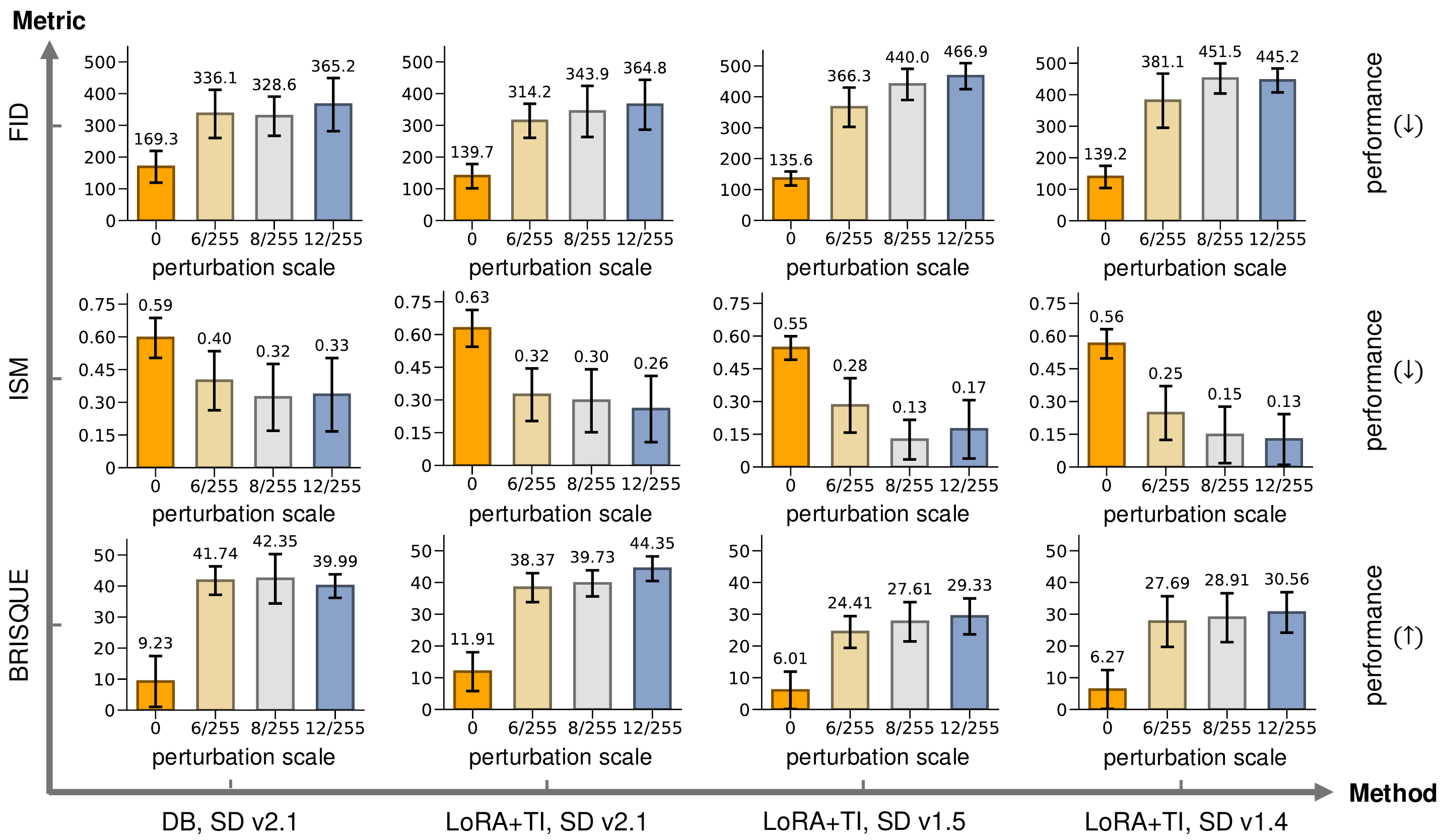}}
\caption{\textbf{Robustness of RID across perturbation levels, pre-trained diffusion models, and personalization techniques.} 
Under the FID, ISM, and BRISQUE metrics, RID demonstrates effective defense performance compared to the baseline on clean images, across three perturbation levels (6/255, 8/255, and 12/255), popular pre-trained diffusion models (SD v2.1, v1.5, and v1.4), and representative personalization techniques (DB and LoRA+TI). In all panels, the $x$-axis represents the perturbation scale, with $0$ indicating baseline results on clean images. Arrows at the end of each row indicate better defense performance (i.e., lower visual quality) under the corresponding metric.
}
\label{fig3}
\end{center}
\vspace{-6mm}
\end{figure*}

To test RID's robustness, we evaluate RID’s defense performance across different perturbation levels, personalization techniques, and pre-trained diffusion models. The results, measured using FID, ISM, and BRISQUE metrics, demonstrate RID’s adaptability and robustness across diverse setups as shown in Fig.~\ref{fig3}.

\paragraph{Robustness to perturbation levels.} The $x$-axes in Fig.~\ref{fig3} represent the perturbation scales (6/255, 8/255, and 12/255), with 0 indicating the baseline results on clean images. As the perturbation scale increases, the defense effectiveness improves, reflected by higher FID and BRISQUE scores and lower ISM scores. Nevertheless, RID provides effective defense even at lower perturbation levels. For example, ISM scores drop from 0.59 to approximately 0.4, while FID increases from 169.26 to over 300, indicating successful disruption of identity retention.

\paragraph{Robustness to personalization techniques.}
We firstly test the robustness of RID over different personalization techniques (aka, DB~\cite{ruiz2023dreambooth} and LoRA+TI~\cite{gal2022image}), based on the pre-trained diffusion model, namely, Stable Diffusion (SD) v2.1. As shown in Fig.~\ref{fig3}, similar trends are observed across both techniques: ISM scores decrease from 0.61 to around 0.3, FID rises from 139.72 to over 300, and BRISQUE increases from 11.91 to 44.35 as the perturbation scale increases. These results demonstrate that RID effectively disrupts identity replication and maintains consistent performance regardless of the specific personalization method.

\paragraph{Robustness to pre-trained diffusion models.}
We further test the robustness of RID across different pre-trained diffusion models, including SD v1.5 and SD v1.4, by employing the LoRA+TI personalization technique. Together with the results of SD v2.1 depicted in Fig.~\ref{fig3}, RID demonstrates consistent trends across all pre-trained diffusion models, with the defense performance improving as the perturbation scale increases. These results further validate RID’s generalizability and effectiveness across diverse model architectures.

Overall, Fig.~\ref{fig3} highlights RID’s ability to deliver robust protection across perturbation levels, 
personalization techniques, and diffusion models. The findings in Fig.~\ref{fig2} and Fig.~\ref{fig3} establish RID as an efficient, effective, and robust solution for safeguarding identities in personalization scenarios.

\subsection*{Analysis of perturbations generated by RID}\label{Sec:ma}

\begin{figure*}[t!]
\begin{center}
\centerline{\includegraphics[width=0.95\linewidth]{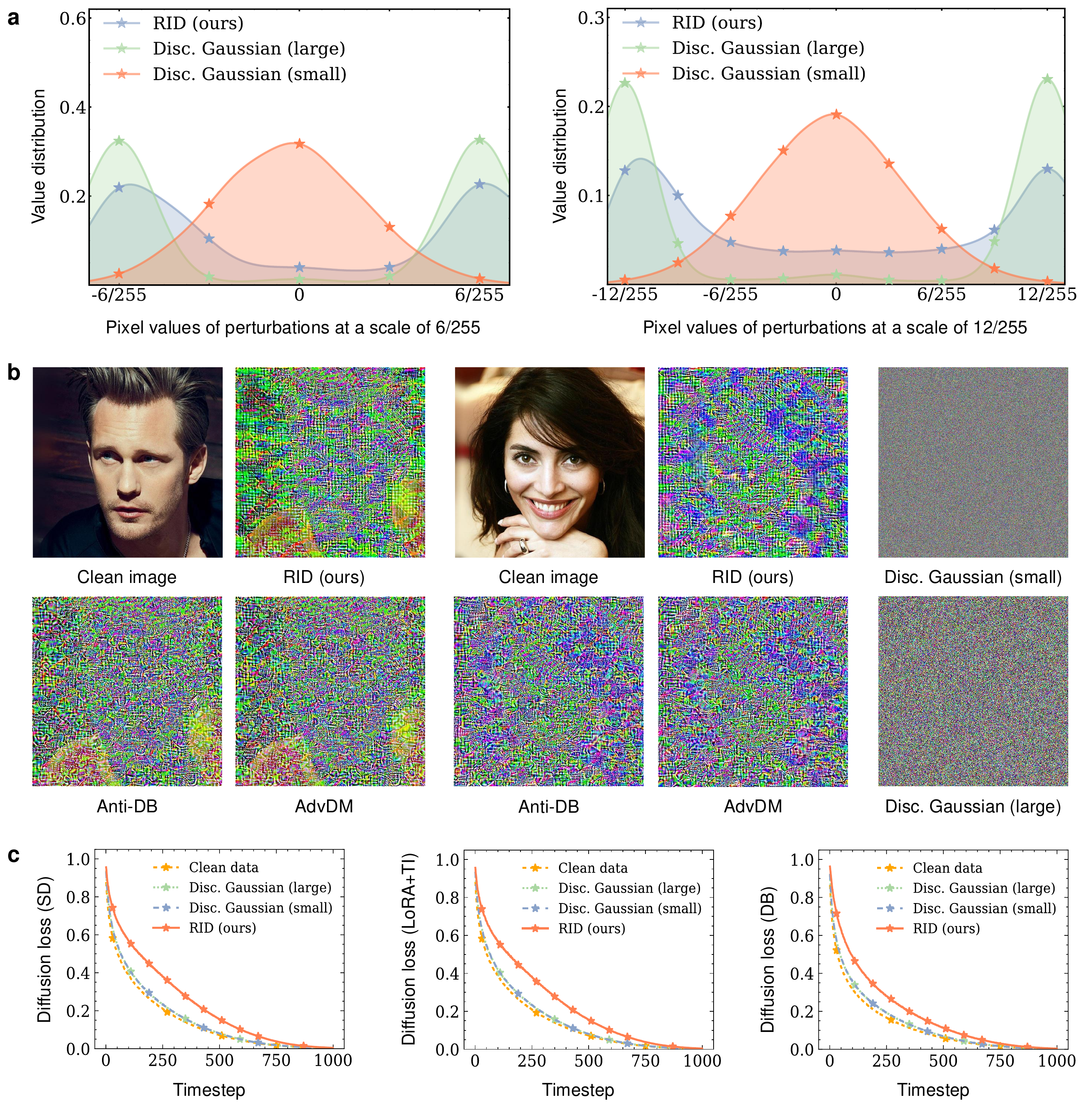}}
\caption{\textbf{Analysis of the perturbations generated by RID.} The pixel values are normalized from $[0, 255]$ to $[-1, 1]$. Two Gaussian noise baselines with standard deviations of the perturbation level divided by $3$ (small, following the 3-sigma rule) and $1$ (large) are used for comparison, discretized to match the perturbation levels and normalized pixel range.
\textbf{a}, Every perturbation is flattened as a vector to count the values of all elements. The distributions of RID perturbations differ markedly from those of Gaussian noise.
\textbf{b}, Visualization of RID perturbations as images reveals structured, semantic patterns aligned with facial features, similar to those of optimization-based methods, while Gaussian noise appears random and unstructured. Perturbations for all methods are displayed at a scale of $1$ (e.g., multiplied by a factor of $255/8$) for clarity.
\textbf{c}, RID perturbations consistently increase diffusion losses compared to Gaussian noise across various pre-trained (SD) and personalized (LoRA+TI and DB) diffusion models at all timesteps, suggesting the effectiveness of learned perturbations over natural noise. All facial images used in this figure are sourced from publicly available datasets and are permitted for academic purposes.
}
\label{fig4}
\end{center}
\vspace{-6mm}
\end{figure*}

This section examines the properties of perturbations generated by RID in comparison with both Gaussian noise baselines and existing optimization-based methods. The corresponding results are reported in Fig.~\ref{fig4}, where RID produces semantically meaningful perturbations that are significantly different from Gaussian noise and behave similarly to those produced by optimization-based methods, such as Anti-DB~\cite{van2023anti} and AdvDM~\cite{liang2023adversarial}.

\paragraph{Pixel distributions of RID perturbations.}
The pixel distributions of RID-generated perturbations, as shown in Fig.~\ref{fig4}\textbf{a}, differ significantly from those of Gaussian noise baselines. All pixel values are normalized to [$-$1, 1], with two Gaussian baselines included for comparison: small Gaussian noise (with a standard deviation of one-third of the perturbation level, following the 3-sigma rule) and large Gaussian noise (matching the full perturbation level). RID perturbations exhibit a heavy-tailed statistical distribution, distinct from the symmetric distribution of Gaussian noise (even with truncation). This characteristic highlights RID’s ability to learn structured perturbations that deviate from the random and unstructured patterns seen in Gaussian noise.

\paragraph{Semantic structure in RID perturbations.}
As shown in Fig.~\ref{fig4}\textbf{b}, RID-generated perturbations reveal structured and semantic patterns that align closely with facial features, similar to those produced by optimization-based methods (e.g., Anti-DB and AdvDM). These perturbations focus on regions containing meaningful information, such as facial contours, to effectively disrupt identity replication. In contrast, Gaussian noise lacks any semantic structure, appearing random and uncorrelated with the content of the image. The semantic alignment of RID perturbations reflects the learned nature of its defense mechanism, ensuring that sensitive features are targeted for protection.

\paragraph{Comparison of diffusion losses.}
Fig.~\ref{fig4}\textbf{c} compares the impact of different perturbation methods, aka, Gaussian noise, RID, and optimization-based methods, on the diffusion loss for both pre-trained and personalized models. Gaussian noise causes only a slight and temporary increase in diffusion loss, which diminishes during the personalization process. As a result, it has minimal effect on the quality of personalized outputs. In contrast, RID perturbations lead to a persistent and substantial increase in diffusion loss, comparable to that achieved by optimization-based methods. This effect holds consistently across LoRA-based and DB-based personalization, where RID prevents the models from effectively fine-tuning to replicate identities. These results demonstrate that RID’s learned perturbations effectively disrupt personalization while offering computational efficiency compared to optimization-based approaches.
   
\begin{figure*}[t!]
\begin{center}
\centerline{\includegraphics[width=0.99\linewidth]{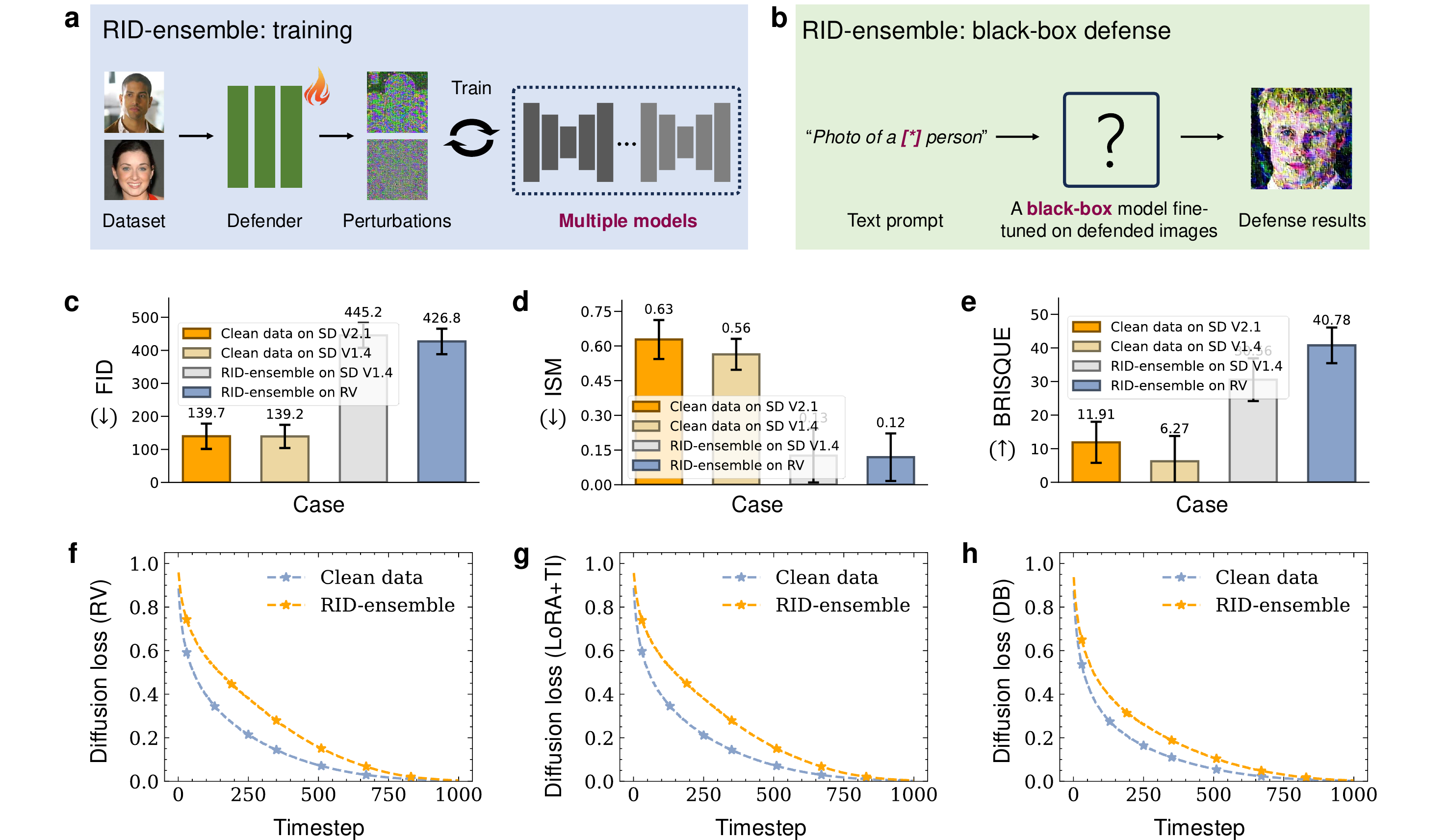}}
\caption{\textbf{RID-Ensemble can successfully defend the identities in black-box settings.}
\textbf{a}, The RID-Ensemble is trained using multiple pre-trained models to enhance its ability to defend identities across models.
\textbf{b}, In a black-box defense scenario, personalized models are trained on defended images using previously unseen pre-trained models. The RID-Ensemble effectively protects identities even against these unknown models.
\textbf{c}, \textbf{d}, \textbf{e}, Quantitative evaluation of RID-Ensemble’s performance shows robust identity protection in black-box defense, as measured by FID, ISM, and BRISQUE.
\textbf{f}, \textbf{g}, \textbf{h},  The effectiveness of RID-Ensemble is further supported by increased diffusion loss across black-box models, indicating strong defense performance in preventing identity retention. All facial images used in this figure are sourced from publicly available datasets and are permitted for academic purposes.
}
\label{fig5}
\end{center}
\vspace{-6mm}
\end{figure*}

\begin{figure*}[t!]
\begin{center}
\centerline{\includegraphics[width=0.96\linewidth]{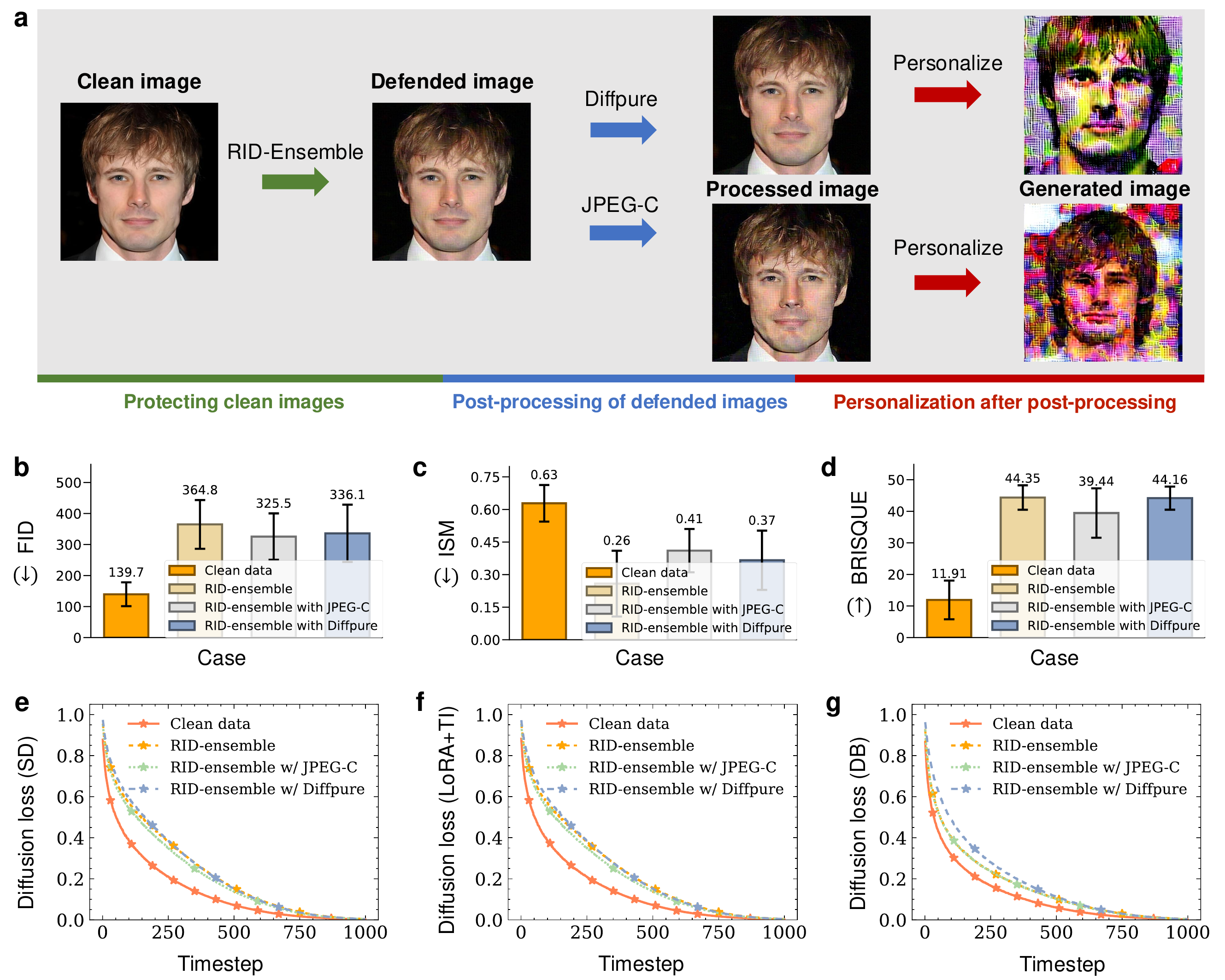}}
\caption{\textbf{RID-Ensemble is robust to post-processing for defended images.}
\textbf{a}, Visual examples show that RID-Ensemble effectively defends images even after post-processing (e.g., JPEG compression and Diffpure).
\textbf{b}, \textbf{c}, \textbf{d}, Quantitative metrics (FID, ISM, and BRISQUE) confirm that RID-Ensemble remains effective under post-processing, particularly when compared to the undefended baseline.
\textbf{e}, \textbf{f}, \textbf{g}, Despite the impact of post-processing, RID-Ensemble maintains higher diffusion losses than undefended images, demonstrating robust defense performance, consistent with the results depicted in \textbf{b}, \textbf{c}, and \textbf{d}, respectively. All facial images used in this figure are sourced from publicly available datasets and are permitted for academic purposes.
}
\label{fig6}
\end{center}
\vspace{-6mm}
\end{figure*}

\subsection*{RID-Ensemble: extensions to black-box and post-processing settings}

For a protection method to be practical, it must demonstrate effectiveness and robustness across diverse scenarios. In real-world applications, two critical challenges arise:
\begin{enumerate}
    \item[(1)] \emph{Black-box defense}: Adversaries may use pre-trained diffusion models not encountered during defender training for personalization.
    \item[(2)] \emph{Post-processing}: Adversaries may apply post-processing techniques to defended images to reduce their protection effectiveness.
\end{enumerate}
To address these challenges, we developed the \emph{RID-Ensemble} inspired by ensemble adversarial training~\cite{tramer2017ensemble}, which utilizes multiple pre-trained diffusion models during training, optimizing the average of their corresponding Adv-SDS losses to ensure generalized robustness (Fig.~\ref{fig4}\textbf{a}).

\paragraph{Black-box protection.} Fig.~\ref{fig5} illustrates the ability of RID-Ensemble to defend identities in black-box scenarios. After training RID-Ensemble using multiple pre-trained models, we tested its robustness against an unseen pre-trained diffusion model, namely, Realistic Vision (RV)~\cite{rv}, used for personalization.

In Fig.~\ref{fig5}\textbf{b}, the qualitative results demonstrate that even when adversaries fine-tune RV on RID-Ensemble-defended images, the generated outputs fail to replicate identities, showing distorted and unrecognizable features. This highlights RID-Ensemble’s generalizability to unseen models.

The quantitative evaluations in Fig.~\ref{fig5}\textbf{c}--\textbf{e} further validate RID-Ensemble’s robust black-box performance. Compared to clean data, defended images achieve higher FID and BRISQUE scores and lower ISM scores, indicating stronger protection against identity retention. Importantly, RID-Ensemble consistently maintains these metrics across both seen and unseen models, demonstrating its adaptability.

Fig.~\ref{fig5}\textbf{f}--\textbf{h} reveals the underlying defense mechanism. Defended images exhibit significantly higher diffusion loss on black-box models compared to clean data, and this loss remains elevated during personalization. The inability of black-box models to mitigate this loss underscores RID-Ensemble’s effectiveness in preventing identity replication, even when models outside its training set are used.

\paragraph{Robustness to post-processing.}  Fig.~\ref{fig6} presents RID-Ensemble’s robustness to post-processing techniques that adversaries may employ to weaken protection. Two representative post-processing methods are analyzed: (1) JPEG compression (JPEG-C)~\cite{guo2017countering}, a traditional approach that compresses high-frequency image information, potentially diminishing the effectiveness of added perturbations; (2) Diffpure~\cite{nie2022diffusion}, a diffusion-based method that applies noise and then denoises defended images, leveraging the generative capacity of diffusion models to restore clean features.

As shown in Fig.~\ref{fig6}\textbf{a}, RID-Ensemble-defended images retain anti-personalization effectiveness even after post-processing. While JPEG-C removes some high-frequency information, RID’s perturbations also contain low-frequency components, ensuring residual defense remains intact. For Diffpure, the high diffusion loss of RID-defended images disrupts the pre-trained model’s perception of image features, further degrading outputs instead of restoring clean identities.

Fig.~\ref{fig6}\textbf{b}--\textbf{d} confirms quantitatively RID-Ensemble’s post-processing robustness. After applying JPEG-C and Diffpure, defended images continue to achieve higher FID and BRISQUE scores and lower ISM scores compared to clean data, indicating persistent protection. Although post-processing reduces defense performance slightly, RID-Ensemble-defended images still maintain strong robustness against identity replication.

Fig.~\ref{fig6}\textbf{e}--\textbf{g} analyzes diffusion losses under post-processing. While post-processing slightly decreases diffusion loss by removing some defending information, RID-Ensemble-defended images still exhibit significantly higher diffusion losses than clean data. As personalization progresses, losses for post-processed images converge to elevated levels, indicating that residual defending information remains effective in disrupting identity replication. 

The results in Figs.~\ref{fig5} and \ref{fig6} validate RID-Ensemble’s effectiveness and robustness in real-world scenarios. These findings demonstrate RID-Ensemble as a practical and reliable solution for safeguarding identities in generative personalization environments.

\section*{Discussion}

Diffusion models, particularly large-scale pre-trained ones, represent a major breakthrough in generative artificial intelligence (AI), rapidly transforming creative industries and gaining an ever-expanding user base. However, alongside their widespread adoption, significant security concerns have emerged. Personalization techniques, while enabling remarkable customization, can be exploited for malicious purposes such as identity fraud and unauthorized image replication. Despite the urgency of these risks, existing methods for identity defense remain impractical due to high computational demands, limiting their application in real-world scenarios.

This paper introduces a novel solution called RID that addresses these challenges by enabling efficient and effective identity defense. Leveraging adversarial score distillation sampling (Adv-SDS), RID generates perturbations with a single network forward pass, providing real-time protection with minimal computational overhead. This remarkable efficiency enables local deployment on user-end devices, such as smartphones, removing the need for cloud-based processing and minimizing the risk of user data leakage. Extensive experiments demonstrate RID’s robust performance in preventing unauthorized identity replication, validating its practicality for real-world use.

While RID represents a significant step forward, several challenges remain. The growing diversity of pre-trained diffusion models poses a notable challenge, as defenders trained on specific models may struggle to generalize to unseen architectures. To address this, the RID-Ensemble framework leverages multiple pre-trained models, enhancing robustness in black-box scenarios. However, RID primarily targets tuning-based personalization methods, such as Dreambooth, which involve fine-tuning model parameters. Emerging tuning-free personalization methods, such as IP-Adapter~\cite{ye2023ip} and InstantID~\cite{wang2024instantid}, introduce unique challenges that require alternative defenses. For tuning-free methods, which leverage encoders rather than fine-tuning, defenses can be applied directly to the encoder, similar to traditional adversarial methods. While tuning-based methods remain the primary focus due to their stronger personalization capabilities, integrating defenses for both paradigms is a promising direction for our future research.

Another critical aspect of identity defense lies in ensuring defended images remain perceptually indistinguishable from clean images. While RID achieves promising protection under relatively small noise constraints, further reducing perturbation visibility is essential for practical deployment in sensitive scenarios. Future work could explore more advanced training techniques and objectives to maximize protection while adhering to stricter imperceptibility requirements. Achieving this balance is crucial for widespread adoption of identity defense technologies.

Going forward, several research avenues hold the potential to enhance the field of identity defense. One promising avenue is optimizing the search space for perturbations. Current methods operate in the pixel domain, but exploring the frequency domain could enable stronger defenses under smaller noise constraints. Additionally, defended perturbations could be transformed into meaningful signals, such as makeup patterns, achieving identity defense while simultaneously adding aesthetic effects. This shift from defensive perturbations to functional enhancements represents an exciting opportunity to expand the utility of these methods.

While our current work focuses on protecting human portraits, intellectual property rights for other creative outputs, such as artwork, also warrant attention. Future research could explore the extension of such a method to prevent personalization techniques from replicating the unique styles of artists without their consent. Robust defenses for artistic works would not only safeguard intellectual property but also contribute to the broader ethical use of generative AI systems.

By addressing the aforementioned challenges and pursuing these future directions, the field of identity defense can evolve into a practical, versatile, and ethically responsible discipline. Ensuring that generative AI systems are harnessed safely will require continued innovation, balancing technical advancements with the protection of personal and creative rights.

\section*{Methods}\label{sec-m}

\subsection*{Background}
\label{subsec-msds0}

Mathematically, a diffusion model~\cite{sohl2015deep,ho2020denoising,song2020score} introduces a forward process that gradually adds noise to data as follows: 
\begin{align} 
x_t = \alpha(t) x_0 + \sigma(t) \epsilon, 
\end{align} 
where $t \in [0, 1]$ represents time, $x_0$ is a randomly selected training image, $\epsilon$ is standard Gaussian noise, and $\alpha(t)$ and $\sigma(t)$ are strictly positive scalar functions of $t$. These functions satisfy that $\alpha(t)/ \sigma(t)$ strictly decreases over $t$, with $\alpha(0)/ \sigma(0)$ being sufficiently large and $\alpha(1)/ \sigma(1)$ sufficiently small. 

The objective of a diffusion model is to reverse the forward process by predicting the noise added to the data at different time steps, which is formalized as follows: 
\begin{align}\label{eq-diff-loss}
    \mathcal{L}_{\text{Diff}}(\theta, x_0)  =  \mathbb{E}_{t,\epsilon} \left [w(t) \left \| \epsilon_\theta(x_t,t)-\epsilon \right \|_2^2 \right],
\end{align}
where $w(t)$ is a strictly positive weighting function over $t$, $\| \cdot \|_2$ denotes the $l_2$ norm, and $\epsilon_{\theta}(\cdot)$ is a noise prediction network parameterized by $\theta$. For simplicity, conditions (e.g., text prompts) are omitted in Eq.~\eqref{eq-diff-loss}. After training, samples can be generated by starting from Gaussian noise and gradually denoising through the noise prediction network.

Personalization methods~\cite{ruiz2023dreambooth,gal2022image} also fine-tune pre-trained diffusion models according to the diffusion loss in Eq.~\eqref{eq-diff-loss} (or its variants) on the customized dataset. Therefore, existing defense methods\cite{van2023anti,wang2023simac} optimize image-specific perturbations to counteract $\mathcal{L}_{\text{Diff}}$, by solving the following optimization problem:
\begin{align}
    \min_{\delta, \|\delta\|_{\infty} \le \varepsilon} -  \mathcal{L}_{\text{Diff}}(\theta, x_0 + \delta),
\end{align}
where $x_0$ is a clean image, $\delta$ is the perturbation to be optimized, $\varepsilon$ is the perturbation level (e.g., $12/255$), $\| \cdot\|_{\infty}$ is the infinity norm and $\theta$ is frozen. For instance, the update rules of AdvDM~\cite{liang2023adversarial} and Anti-DB~\cite{van2023anti} are given by:
\begin{subequations}
\begin{align}\label{eq-antidb}
     \delta & \leftarrow \delta + \gamma \text{sign}( \nabla_{\delta} \mathcal{L}_{\text{Diff}}(\theta, x_0^{\textrm{def}})), \\
   \delta & \leftarrow \text{clamp}(\delta,-\varepsilon,\varepsilon ) ,
\end{align}
\end{subequations}
where $\leftarrow$ represents the assignment operation, $\gamma$ is the step size, typically set to $1/10$ of the noise constraint $\varepsilon$, $x_0^{\textrm{def}} \triangleq x_0+ \delta$ is the the corresponding defended image and $\text{sign}(\cdot)$ denotes the sign function, which outputs values of $-1$ or $1$. The function $\text{clamp}(\cdot)$ clips the values of $\delta$ to lie within the range $[-\varepsilon, \varepsilon]$. Here, $\theta$ corresponds to the pre-trained diffusion models in AdvDM~\cite{liang2023adversarial} and surrogate personalized models in Anti-DB~\cite{van2023anti}, respectively.  

\subsection*{Adversarial score distillation sampling} 

In contrast to existing defense methods, the core principle of RID lies in training a neural network to generate protective perturbations for each image via a forward inference pass, eliminating the need for individual optimization. Inspired by Differentiable Image Parameterization (DIP)~\cite{mordvintsev2018differentiable}, RID represents the defended image and its noisy versions using differentiable parameters below:
\begin{subequations}
\begin{align}
\label{eq-rid-x0}
    x_0^\text{def} & = x_0+ \delta_{\phi}(x_0), \\ x_t^{\text{def}} & = \alpha(t) x_0^{\text{def}} +\sigma(t) \epsilon, \label{eq-rid-xt}
\end{align}
\end{subequations}
where $\delta_{\phi} (\cdot)$ is the neural network in RID parameterized by $\phi$, and $\alpha(t)$ and $\sigma(t)$ are the same scalar functions used in the pre-trained diffusion model. The network $\delta_{\phi} (\cdot)$ is shared across all images and is trained on a multi-person dataset (as specified later) to generalize to new images.

For the training objective, a natural approach would be to attack the diffusion loss in Eq.~\eqref{eq-diff-loss} by plugging in the defended image from Eq.~\eqref{eq-rid-x0}. Formally, using the chain rule, the gradient of $\phi$ can be expressed as:
\begin{align}\label{eq-diff-grad}
   \nabla_{\phi} \left(- \mathcal{L}_{\text{Diff}}(\phi, x_0) \right) =   -\mathbb{E}_{t,\epsilon}\left[w(t) \alpha(t) (\epsilon_{\theta}(x_t^\text{def},t)-\epsilon) \frac{\partial\epsilon_\theta(x_t^\text{def},t)}{\partial x_t^{\textrm{def}}}   \frac{\partial x_0^{\textrm{def}}}{\partial \phi} \right],
\end{align}
where $x_0$ is a clean image. However, this gradient calculation involves the Jacobian $\partial \epsilon_\theta(x_t^\text{def}, t)/\partial x_t^{\textrm{def}}$ of the pre-trained diffusion model, which is computationally intensive due to the model’s large size. Additionally, it is highly nontrivial to update $\phi$ through backpropagation if non-differentiable operations such as $\textrm{sign}(\cdot)$ and $\textrm{clamp}(\cdot)$ in  Eq.~\eqref{eq-antidb} on $\delta_{\phi} (x_0)$ are introduced.

To address these challenges, we draw inspiration from \emph{Score Distillation Sampling} (SDS) in DreamFusion~\cite{poole2022dreamfusion}, which optimizes 3D structures by distilling information from a diffusion model on images. 
While the original SDS minimizes diffusion loss to improve the quality of 3D asserts, our approach aims to maximize diffusion loss on perturbed images adversarially to defend identities. Therefore, we call this approach \emph{Adversarial Score Distillation Sampling} (Adv-SDS).

We simplify the gradient in Eq.~\eqref{eq-diff-grad} by removing the Jacobian term, resulting in the following update rule for $\phi$ given an image $x_0$: 
\begin{equation}
\label{eq-advsds}
\begin{aligned}
\nabla_{\phi }\mathcal{L}_{\text{Adv-SDS}}(\phi, x_0) & = -\mathbb{E}_{t,\epsilon}\left[w(t) \left(\epsilon_{\theta}(x_t^\text{def})-\epsilon\right)\frac{\partial x_0^{\textrm{def}}}{\partial \phi} \right], \\ 
& = -\mathbb{E}_{t,\epsilon} \left[w(t) \left(\epsilon_{\theta}(x_t^{\textrm{def}})-\epsilon\right)\frac{\partial \delta_\phi(x_0)}{\partial \phi}\right],
\end{aligned}
\end{equation}
where $\theta$ is frozen. Following the proof in DreamFusion~\cite{poole2022dreamfusion}, this update rule effectively minimizes the negative of a weighted KL divergence:
\begin{align} \label{eq-advsds-loss}
    \mathcal{L}_{\text{Adv-SDS}}(\phi, x_0) = -\mathbb{E}_{t} \left[ \sigma(t) / \alpha(t) w(t) \textrm{KL} (q(x^\text{def}_t|x_0^\textrm{def}, t)||p_{\theta}(x_t^\text{def}, t))\right],
\end{align}
where $q(\cdot|\cdot, t)$ is the Gaussian distribution defined by the noise-adding process in Eq.~\eqref{eq-rid-xt} at time $t$, and $p_{\theta}(\cdot, t)$ is the corresponding marginal distribution of the diffusion model’s sampling process. Intuitively, the loss function in Eq.~\eqref{eq-advsds-loss} enforces the learned sampling process to diverge from the forward process on defended images, thereby protecting identities.

However, directly calculating the loss in Eq.~\eqref{eq-advsds-loss} is intractable because $p_{\theta}(\cdot, t)$ is parameterized by the noise prediction network. To address this, we employ a surrogate loss to monitor the optimization process and simplify the implementation:
\begin{align}
\label{eq-suradvsds}
\mathcal{L}_{\text{Sur-Adv-SDS}}(\phi, x_0)  = \mathbb{E}_{t,\epsilon}\left[\left \|  x_t^{\text{def}} - \textrm{stop-grad}\left( x_t^{\text{def}}+w(t)\left(\epsilon_\theta( x_t^{\text{def}},t)- \epsilon \right)\right)\right \|_2^2 \right],
\end{align}
where $\text{stop-grad}(\cdot)$ prevents gradient flow through the enclosed terms.

It can be shown that $\nabla_{\phi }\mathcal{L}_{\text{Sur-Adv-SDS}} \propto \nabla{\phi }\mathcal{L}_{\text{Adv-SDS}}$, making Eq.~\eqref{eq-suradvsds} an equivalent objective for optimizing $\phi$ via stochastic gradient descent. Additionally, the $\text{stop-grad}(\cdot)$ operator does not affect the surrogate loss’s value, which can be simplified to $\mathbb{E}_{t,\epsilon}\left[w^2(t) \left \| \epsilon_\theta(x_t^{\text{def}}, t) - \epsilon \right \|_2^2\right]$, a reweighted version of the diffusion loss in Eq.~\eqref{eq-diff-loss}. 
For simplicity, we use a constant weighting function $w(t)$, which makes the surrogate loss equivalent to the original diffusion loss up to a multiplicative factor. Consequently, this surrogate loss indicates the RID’s training progress: as training proceeds, the expected value of the surrogate loss increases, indicating improved defense performance in theory.

Through experiments, we found that training RID solely with $\mathcal{L}_{\text{Sur-Adv-SDS}}$ achieved a reasonably high level of protection. However, the generated perturbations exhibited subtle grid-like patterns, slightly affecting the images’ visual appearance (see Extended Data Fig.~\ref{fig7}\textbf{a}). To improve visual consistency with the original images, we introduced an additional regularization term defined as: 
\begin{align}
\label{eq:reg}
\mathcal{L}_{\textrm{reg}}(\phi, x_0) = \| \delta_\phi(x_0) - \delta^{\textrm{iw}}(x_0)\|_1,
\end{align}
where $\|\cdot\|_1$ is the $l_1$ norm, which we found to perform better in practice than the $l_2$ norm. Here, $\delta^{\textrm{iw}}(x_0)$ represents the perturbation generated by an image-wise attack method, such as Anti-DB~\cite{van2023anti}.

Our final objective is to optimize
\begin{align}\label{eq-final}
    \mathbb{E}_{x_0\sim \mathcal{D}} \left [ \mathcal{L}_{\textrm{Sur-Adv-SDS}}(\phi, x_0) \right ] + \lambda \mathbb{E}_{x_0\sim \mathcal{D}'} \left [\mathcal{L}_{\textrm{reg}}(\phi, x_0) \right ],
\end{align}
where $\lambda$ is a hyperparameter balancing the two loss functions. We set $\lambda=3$ by default to ensure both losses are of comparable magnitude. Here, $\mathcal{D}$ denotes the full training set, and $\mathcal{D}'$ is a subset of $\mathcal{D}$ as specified later. We trained RID using automatic differentiation in PyTorch according to Eq.~\eqref{eq-final}. In our experiments, the regularization term effectively mitigated the grid-like artifact issue. However, when applied alone, it failed to provide adequate protection—identity replication remained possible, albeit with reduced realism, as shown in Extended Data Fig.~\ref{fig7}\textbf{b}. In contrast, the combined loss function delivers the best qualitative and quantitative performance to ensure robust protection, as demonstrated in Extended Data Fig.~\ref{fig7}\textbf{b}-\textbf{c}.
 
\subsection*{Training data}
 
To ensure robust performance across diverse facial features, we utilized the widely recognized VGG-Face2 dataset~\cite{simonyan2014very}, which contains approximately 3.31 million images spanning 9,131 distinct identities. This dataset primarily features human faces captured under varying poses, lighting conditions, and expressions. For our purposes, we adopted a filtered subset following the methodology described in SimSwap++~\cite{chen2023simswap++} and Anti-Dreambooth (Anti-DB)\cite{van2023anti}. The filtering criteria required each identity to have at least 15 images with a resolution of at least 500 × 500 pixels. This process resulted in a curated training set of 70,000 high-resolution facial images.

We used null text for all images during training, enabling RID to defend images without imposing restrictions on the subsequent personalization of prompts. Through experiments, we found that null text training works well and consistently across different settings.

We prepared pairs of clean images and corresponding perturbations to facilitate RID training. Specifically, we randomly selected 7,000 images (10\% of the filtered dataset) and optimized the corresponding perturbations using Anti-DB~\cite{van2023anti} under default settings. The optimized perturbations, along with their corresponding clean images, were stored as fixed pairs. During training, two data loaders were used: one to import paired data (clean images with their perturbations) and another to handle unpaired data, ensuring flexibility in training dynamics.

\subsection*{Model architecture}\label{subsec-m1}

RID utilizes a variant of the Diffusion Transformer (DiT)~\cite{peebles2023scalable} as its backbone. DiT naturally ensures that the output dimensionality matches the input and has proven effective in noise prediction tasks, which closely align with our objective of perturbation prediction. Unlike the original DiT, our approach does not require additional conditioning inputs, such as class or text conditioning. Instead, these components are replaced with zero vectors of the same dimension. This simplification reduces the model's complexity while preserving its ability to generate effective protective perturbations.

Extended Data Table~\ref{tab:model_specs} summarizes the settings of various RID configurations and their inference costs, measured in Gflops. As the protection performance across different RID configurations is similar, we adopt the fastest by default to minimize computational overhead.

To ensure the perturbations remain imperceptible, we constrain their magnitude using the infinity norm, consistent with existing methods. However, rather than relying on conventional output clipping, we adopt an end-to-end forward process that incorporates a \textbf{scaled tanh} function in the network’s final layer. This ensures the output perturbations lie within $[-\varepsilon, \varepsilon]$. This design enhances optimization stability by preventing zero gradients in clipped dimensions during training.

\subsection*{RID-Ensemble}

To enhance protection capabilities in black-box settings and against post-processing, we extend RID to an ensemble version, termed RID-Ensemble. During training, a single RID-Ensemble is optimized to defend against $N$ pre-trained diffusion models. For each $i = 1, \dots, N$, the forward process of the $i$-th model is defined as:
\begin{subequations}
\begin{align}
\label{eq-ensemble-x0}
    x_0^\text{def-ens} & = x_0+ \delta_{\phi_\text{ens}}(x_0), \\
    x_t^{\text{def},(i)}  &= \alpha_i(t) x_0^{\text{def-ens}} +\sigma_i(t) \epsilon, 
\end{align}
\end{subequations}
where $\alpha_i(t)$ and $\sigma_i(t)$ are scalar functions specific to the $i$-th pre-trained diffusion model. The neural network $\delta_{\phi_\text{ens}} (\cdot)$, parameterized by $\phi_\text{ens}$, is shared across all pre-trained models, enabling RID-Ensemble to defend against multiple diffusion models with a single network.
 
To train RID-Ensemble, following Eq.~\eqref{eq-suradvsds}, we optimize the surrogate loss for the $i$-th pre-trained diffusion model as: 
\begin{align}
\label{eq-suradvsds-ens}
\mathcal{L}_{\text{Sur-Adv-SDS-}(i)}(\phi_\text{ens}, x_0)  = \mathbb{E}_{t,\epsilon}\left[\left \|  x_t^{\text{def-ens}} - \textrm{stop-grad}\left( x_t^{\text{def},(i)}+w(t)\left(\epsilon_{\theta_{(i)}}( x_t^{\text{def},(i)},t)- \epsilon \right)\right)\right \|_2^2 \right],
\end{align}
where $\epsilon_{\theta_{(i)}}$ represents the noise prediction network of the $i$-th pre-trained model, and $\theta_{(i)}$ is frozen during training. The final objective for training RID-Ensemble is given by:
\begin{align}\label{eq-final-ens}
    \frac{1}{N}\sum_{i=1}^N\mathbb{E}_{x_0\sim \mathcal{D}} \left [ \mathcal{L}_{\textrm{Sur-Adv-SDS-}(i)}(\phi_{\text{ens}}, x_0) \right ] + \lambda \mathbb{E}_{x_0\sim \mathcal{D}'} \left [\mathcal{L}_{\textrm{reg}}(\phi_{\text{ens}}, x_0) \right ],
\end{align}
where the training datasets $\mathcal{D}$ and $\mathcal{D}'$ as well as the value of $\lambda$ are the same as in Eq.~\eqref{eq-final}.  In our default setting, we have $N=3$ with SD v1.4, SD v1.5 and SD v2.1.

\subsection*{Evaluation metrics}
\label{supp:metrics}

In this section, we present the metrics used to evaluate the protection performance.

To evaluate image quality, we utilize the widely-adopted Frechet Inception Distance (FID) metric~\cite{heusel2017gans}, which has become a standard benchmark in image generation tasks. The FID metric operates by first extracting features using the Inception network~\cite{szegedy2016rethinking}, then modeling the feature distribution with a multivariate Gaussian model. The distance between the generated and reference samples is subsequently calculated based on these feature distributions as follows:
\begin{align}
\textrm{FID}=  \| \mu_g-\mu_r  \|_2^2 + \textrm{tr}\left(\Sigma_g+\Sigma_r-2(\Sigma_g \Sigma_r)^{1/2}\right),
\end{align}
where $\mu_g$ and $\mu_r$ represent the mean feature vectors of the generated samples and reference samples, respectively, and $\Sigma_g$ and $\Sigma_r$ the covariance matrices of the features for the generated and reference samples. $\textrm{tr}$ refers to the trace operator. A lower FID score indicates a smaller distance between the generated and real image distributions, corresponding to more realistic generated samples.

To evaluate the similarity between the generated images and the reference images, we employ the Identity Score Matching (ISM) metric~\cite{van2023anti}. ISM utilizes ArcFace~\cite{deng2019arcface}, a strong recognition model, to detect and extract face embeddings from both the generated and reference images. The similarity score is computed as:
\begin{align}
\textrm{ISM} = \frac{\bar{\textbf{e}}_{g}\cdot \bar{\textbf{e}}_{r} }{\left \| \bar{\textbf{e}}_{g} \right \|_2 \left \| \bar{\textbf{e}}_{r} \right \|_2 } ,
\end{align}
where $\bar{\textbf{e}}_{g}$ and $\bar{\textbf{e}}_{r}$ represent the mean face embeddings extracted from the generated images and reference images, respectively. However, since some generated images may be of insufficient quality for successful face detection (note that this is the purpose of identity protection), we introduce an adjusted ISM metric that accounts for the face detection success rate and is defined as $\textrm{aISM} = \textrm{ISM} \cdot \textrm{DR}$, where $\textrm{DR}$ denotes the success ratio of face detection. This adjustment ensures that the similarity score reflects both the quality of the generated faces and the robustness of face detection.

BRISQUE~\cite{mittal2012no} is a reference-free image quality assessment method, where higher scores indicate lower perceptual quality. It evaluates image quality by analyzing natural scene statistics in the spatial domain. First, it computes Mean Subtracted Contrast Normalized (MSCN) coefficients:
\begin{equation}
\hat{I}(i,j) = \frac{I(i,j) - \mu(i,j)}{\sigma(i,j) + C},
\end{equation}
where \( I(i,j) \) is the pixel intensity, \( \mu(i,j) \) and \( \sigma(i,j) \) are the local mean and variance, respectively, and \( C \) is a small constant to avoid division by zero. 
The local mean and variance are computed using a Gaussian kernel $W$ as follows
\begin{equation}
\mu = W * I, \quad \sigma = \sqrt{W * (I - \mu)^2},
\end{equation}
where $*$ denotes the convolution operator. Next, pairwise products of MSCN coefficients are computed in four orientations: horizontal (H), vertical (V), left diagonal (D1), and right diagonal (D2):
\begin{subequations}
\begin{align}
H(i, j) & = \hat{I}(i, j) \cdot \hat{I}(i, j+1), \\
V(i, j) & = \hat{I}(i, j) \cdot \hat{I}(i+1, j), \\
D1(i, j) & = \hat{I}(i, j) \cdot \hat{I}(i+1, j+1), \\
D2(i, j) & = \hat{I}(i, j) \cdot \hat{I}(i+1, j-1).
\end{align}
\end{subequations}
The MSCN coefficients and pairwise product images are fitted to the generalized Gaussian distribution and asymmetric generalized Gaussian distribution, respectively, to extract shape and variance parameters. These features are used to train a classifier that predicts perceptual quality. Lower BRISQUE scores indicate higher image quality. In the context of defended images, increased BRISQUE scores reflect perceptual degradation, indicating a successful defense.

\section*{Data availability} 
All facial data used in this study were obtained from two publicly available and widely used datasets for machine learning. The licenses associated with these datasets permit their use for non-commercial academic purposes. The datasets and licenses are accessible on GitHub at \url{https://github.com/CelebV-HQ/CelebV-HQ} and \url{https://github.com/cydonia999/VGGFace2-pytorch}.

\section*{Code availability} 
All the source codes to reproduce the results in this study are available on GitHub at \url{https://github.com/Guohanzhong/RID}.

\bibliographystyle{unsrt}
\bibliography{references}

\begin{thebibliography}{10}

\bibitem{ramesh2021zero}
Aditya Ramesh, Mikhail Pavlov, Gabriel Goh, Scott Gray, Chelsea Voss, Alec Radford, Mark Chen, and Ilya Sutskever.
\newblock Zero-shot text-to-image generation.
\newblock In {\em International conference on machine learning}, pages 8821--8831. Pmlr, 2021.

\bibitem{rombach2022high}
Robin Rombach, Andreas Blattmann, Dominik Lorenz, Patrick Esser, and Bj{\"o}rn Ommer.
\newblock High-resolution image synthesis with latent diffusion models.
\newblock In {\em Proceedings of the IEEE/CVF conference on computer vision and pattern recognition}, pages 10684--10695, 2022.

\bibitem{saharia2022photorealistic}
Chitwan Saharia, William Chan, Saurabh Saxena, Lala Li, Jay Whang, Emily~L Denton, Kamyar Ghasemipour, Raphael Gontijo~Lopes, Burcu Karagol~Ayan, Tim Salimans, et~al.
\newblock Photorealistic text-to-image diffusion models with deep language understanding.
\newblock {\em Advances in neural information processing systems}, 35:36479--36494, 2022.

\bibitem{betker2023improving}
James Betker, Gabriel Goh, Li~Jing, Tim Brooks, Jianfeng Wang, Linjie Li, Long Ouyang, Juntang Zhuang, Joyce Lee, Yufei Guo, et~al.
\newblock Improving image generation with better captions.
\newblock {\em Computer Science. https://cdn. openai. com/papers/dall-e-3. pdf}, 2(3):8, 2023.

\bibitem{podellsdxl}
Dustin Podell, Zion English, Kyle Lacey, Andreas Blattmann, Tim Dockhorn, Jonas M{\"u}ller, Joe Penna, and Robin Rombach.
\newblock Sdxl: Improving latent diffusion models for high-resolution image synthesis.
\newblock In {\em The Twelfth International Conference on Learning Representations}, 2024.

\bibitem{esser2024scaling}
Patrick Esser, Sumith Kulal, Andreas Blattmann, Rahim Entezari, Jonas M{\"u}ller, Harry Saini, Yam Levi, Dominik Lorenz, Axel Sauer, Frederic Boesel, et~al.
\newblock Scaling rectified flow transformers for high-resolution image synthesis.
\newblock In {\em Forty-first International Conference on Machine Learning}, 2024.

\bibitem{kumari2023multi}
Nupur Kumari, Bingliang Zhang, Richard Zhang, Eli Shechtman, and Jun-Yan Zhu.
\newblock Multi-concept customization of text-to-image diffusion.
\newblock In {\em Proceedings of the IEEE/CVF Conference on Computer Vision and Pattern Recognition}, pages 1931--1941, 2023.

\bibitem{dong2022dreamartist}
Ziyi Dong, Pengxu Wei, and Liang Lin.
\newblock Dreamartist: Towards controllable one-shot text-to-image generation via contrastive prompt-tuning.
\newblock {\em arXiv preprint arXiv:2211.11337}, 2022.

\bibitem{han2023svdiff}
Ligong Han, Yinxiao Li, Han Zhang, Peyman Milanfar, Dimitris Metaxas, and Feng Yang.
\newblock Svdiff: Compact parameter space for diffusion fine-tuning.
\newblock In {\em Proceedings of the IEEE/CVF International Conference on Computer Vision}, pages 7323--7334, 2023.

\bibitem{xiang2023closer}
Chendong Xiang, Fan Bao, Chongxuan Li, Hang Su, and Jun Zhu.
\newblock A closer look at parameter-efficient tuning in diffusion models.
\newblock {\em arXiv preprint arXiv:2303.18181}, 2023.

\bibitem{xiao2023fastcomposer}
Guangxuan Xiao, Tianwei Yin, William~T Freeman, Fr{\'e}do Durand, and Song Han.
\newblock Fastcomposer: Tuning-free multi-subject image generation with localized attention.
\newblock {\em International Journal of Computer Vision}, pages 1--20, 2024.

\bibitem{liu2023facechain}
Yang Liu, Cheng Yu, Lei Shang, Ziheng Wu, Xingjun Wang, Yuze Zhao, Lin Zhu, Chen Cheng, Weitao Chen, Chao Xu, et~al.
\newblock Facechain: A playground for identity-preserving portrait generation.
\newblock {\em arXiv preprint arXiv:2308.14256}, 2023.

\bibitem{ding2022delta}
Ning Ding, Yujia Qin, Guang Yang, Fuchao Wei, Zonghan Yang, Yusheng Su, Shengding Hu, Yulin Chen, Chi-Min Chan, Weize Chen, et~al.
\newblock Parameter-efficient fine-tuning of large-scale pre-trained language models.
\newblock {\em Nature Machine Intelligence}, 5(3):220--235, 2023.

\bibitem{sd15}
Stability, 2023.
\newblock \url{https://huggingface.co/runwayml/stable-diffusion-v1-5}.

\bibitem{sd21}
Stability, 2023.
\newblock \url{https://huggingface.co/stabilityai/stable-diffusion-2-1-base}.

\bibitem{shan2023glaze}
Shawn Shan, Jenna Cryan, Emily Wenger, Haitao Zheng, Rana Hanocka, and Ben~Y Zhao.
\newblock Glaze: Protecting artists from style mimicry by $\{$Text-to-Image$\}$ models.
\newblock In {\em 32nd USENIX Security Symposium (USENIX Security 23)}, pages 2187--2204, 2023.

\bibitem{liang2023adversarial}
Chumeng Liang, Xiaoyu Wu, Yang Hua, Jiaru Zhang, Yiming Xue, Tao Song, Zhengui Xue, Ruhui Ma, and Haibing Guan.
\newblock Adversarial example does good: Preventing painting imitation from diffusion models via adversarial examples.
\newblock In {\em International Conference on Machine Learning}, pages 20763--20786. PMLR, 2023.

\bibitem{madry2017towards}
Aleksander Madry, Aleksandar Makelov, Ludwig Schmidt, Dimitris Tsipras, and Adrian Vladu.
\newblock Towards deep learning models resistant to adversarial attacks.
\newblock {\em arXiv preprint arXiv:1706.06083}, 2017.

\bibitem{van2023anti}
Thanh Van~Le, Hao Phung, Thuan~Hoang Nguyen, Quan Dao, Ngoc~N Tran, and Anh Tran.
\newblock Anti-dreambooth: Protecting users from personalized text-to-image synthesis.
\newblock In {\em Proceedings of the IEEE/CVF International Conference on Computer Vision}, pages 2116--2127, 2023.

\bibitem{wang2023simac}
Feifei Wang, Zhentao Tan, Tianyi Wei, Yue Wu, and Qidong Huang.
\newblock Simac: A simple anti-customization method against text-to-image synthesis of diffusion models.
\newblock {\em arXiv preprint arXiv:2312.07865}, 2023.

\bibitem{salman2023raising}
Hadi Salman, Alaa Khaddaj, Guillaume Leclerc, Andrew Ilyas, and Aleksander M{\k{a}}dry.
\newblock Raising the cost of malicious ai-powered image editing.
\newblock In {\em Proceedings of the 40th International Conference on Machine Learning}, pages 29894--29918, 2023.

\bibitem{goodfellow2015explaining}
Ian~J Goodfellow, Jonathon Shlens, and Christian Szegedy.
\newblock Explaining and harnessing adversarial examples.
\newblock {\em stat}, 1050:20, 2015.

\bibitem{ho2020denoising}
Jonathan Ho, Ajay Jain, and Pieter Abbeel.
\newblock Denoising diffusion probabilistic models.
\newblock {\em Advances in Neural Information Processing Systems}, 33:6840--6851, 2020.

\bibitem{poole2022dreamfusion}
Ben Poole, Ajay Jain, Jonathan~T Barron, and Ben Mildenhall.
\newblock Dreamfusion: Text-to-3d using 2d diffusion.
\newblock In {\em The Eleventh International Conference on Learning Representations}, 2023.

\bibitem{simonyan2014very}
Karen Simonyan and Andrew Zisserman.
\newblock Very deep convolutional networks for large-scale image recognition.
\newblock In {\em The Third International Conference on Learning Representations}, 2015.

\bibitem{chen2023simswap++}
Xuanhong Chen, Bingbing Ni, Yutian Liu, Naiyuan Liu, Zhilin Zeng, and Hang Wang.
\newblock Simswap++: Towards faster and high-quality identity swapping.
\newblock {\em IEEE Transactions on Pattern Analysis and Machine Intelligence}, 2023.

\bibitem{karras2017progressive}
Tero Karras, Timo Aila, Samuli Laine, and Jaakko Lehtinen.
\newblock Progressive growing of gans for improved quality, stability, and variation.
\newblock In {\em International Conference on Learning Representations}, 2018.

\bibitem{ruiz2023dreambooth}
Nataniel Ruiz, Yuanzhen Li, Varun Jampani, Yael Pritch, Michael Rubinstein, and Kfir Aberman.
\newblock Dreambooth: Fine tuning text-to-image diffusion models for subject-driven generation.
\newblock In {\em Proceedings of the IEEE/CVF Conference on Computer Vision and Pattern Recognition}, pages 22500--22510, 2023.

\bibitem{gal2022image}
Rinon Gal, Yuval Alaluf, Yuval Atzmon, Or~Patashnik, Amit~H Bermano, Gal Chechik, and Daniel Cohen-Or.
\newblock An image is worth one word: Personalizing text-to-image generation using textual inversion.
\newblock In {\em The Eleventh International Conference on Learning Representations}, 2023.

\bibitem{hu2021lora}
Edward~J Hu, Yelong Shen, Phillip Wallis, Zeyuan Allen-Zhu, Yuanzhi Li, Shean Wang, Lu~Wang, and Weizhu Chen.
\newblock Lora: Low-rank adaptation of large language models.
\newblock In {\em The Tenth International Conference on Learning Representations}, 2022.

\bibitem{heusel2017gans}
Martin Heusel, Hubert Ramsauer, Thomas Unterthiner, Bernhard Nessler, and Sepp Hochreiter.
\newblock Gans trained by a two time-scale update rule converge to a local nash equilibrium.
\newblock {\em Advances in neural information processing systems}, 30, 2017.

\bibitem{deng2019arcface}
Jiankang Deng, Jia Guo, Niannan Xue, and Stefanos Zafeiriou.
\newblock Arcface: Additive angular margin loss for deep face recognition.
\newblock In {\em Proceedings of the IEEE/CVF conference on computer vision and pattern recognition}, pages 4690--4699, 2019.

\bibitem{mittal2012no}
Anish Mittal, Anush~Krishna Moorthy, and Alan~Conrad Bovik.
\newblock No-reference image quality assessment in the spatial domain.
\newblock {\em IEEE Transactions on image processing}, 21(12):4695--4708, 2012.

\bibitem{tramer2017ensemble}
Florian Tram{\`e}r, Alexey Kurakin, Nicolas Papernot, Ian Goodfellow, Dan Boneh, and Patrick McDaniel.
\newblock Ensemble adversarial training: Attacks and defenses.
\newblock In {\em International Conference on Learning Representations}, 2018.

\bibitem{rv}
Realistic Vision, 2023.
\newblock \url{https://civitai.com/models/4201?modelVersionId=501240}.

\bibitem{guo2017countering}
Chuan Guo, Mayank Rana, Moustapha Cisse, and Laurens van~der Maaten.
\newblock Countering adversarial images using input transformations.
\newblock In {\em International Conference on Learning Representations}, 2018.

\bibitem{nie2022diffusion}
Weili Nie, Brandon Guo, Yujia Huang, Chaowei Xiao, Arash Vahdat, and Animashree Anandkumar.
\newblock Diffusion models for adversarial purification.
\newblock In {\em International Conference on Machine Learning}, pages 16805--16827. PMLR, 2022.

\bibitem{ye2023ip}
Hu~Ye, Jun Zhang, Sibo Liu, Xiao Han, and Wei Yang.
\newblock Ip-adapter: Text compatible image prompt adapter for text-to-image diffusion models.
\newblock {\em arXiv preprint arXiv:2308.06721}, 2023.

\bibitem{wang2024instantid}
Qixun Wang, Xu~Bai, Haofan Wang, Zekui Qin, and Anthony Chen.
\newblock Instantid: Zero-shot identity-preserving generation in seconds.
\newblock {\em arXiv preprint arXiv:2401.07519}, 2024.

\bibitem{sohl2015deep}
Jascha Sohl-Dickstein, Eric Weiss, Niru Maheswaranathan, and Surya Ganguli.
\newblock Deep unsupervised learning using nonequilibrium thermodynamics.
\newblock In {\em International conference on machine learning}, pages 2256--2265. PMLR, 2015.

\bibitem{song2020score}
Yang Song, Jascha Sohl-Dickstein, Diederik~P Kingma, Abhishek Kumar, Stefano Ermon, and Ben Poole.
\newblock Score-based generative modeling through stochastic differential equations.
\newblock In {\em The Nineth International Conference on Learning Representations}, 2021.

\bibitem{mordvintsev2018differentiable}
Alexander Mordvintsev, Nicola Pezzotti, Ludwig Schubert, and Chris Olah.
\newblock Differentiable image parameterizations.
\newblock {\em Distill}, 3(7):e12, 2018.

\bibitem{peebles2023scalable}
William Peebles and Saining Xie.
\newblock Scalable diffusion models with transformers.
\newblock In {\em Proceedings of the IEEE/CVF International Conference on Computer Vision}, pages 4195--4205, 2023.

\bibitem{szegedy2016rethinking}
Christian Szegedy, Vincent Vanhoucke, Sergey Ioffe, Jon Shlens, and Zbigniew Wojna.
\newblock Rethinking the inception architecture for computer vision.
\newblock In {\em Proceedings of the IEEE conference on computer vision and pattern recognition}, pages 2818--2826, 2016.

\end{thebibliography}

\vspace{12pt}
\section*{Acknowledgement}
 
This work was supported by the National Natural Science Foundation of China (No. 92470118, No. 92270118, No. 62276269),  Beijing Natural Science Foundation (No. L247030, No. 1232009), and Beijing Nova Program (No. 20220484044). The work was partially done at the Engineering Research Center of Next-Generation Intelligent Search and Recommendation, Ministry of Education.

\section*{Author contributions} 

H.G., S.N., and C.L. proposed the initial idea. H.S. and C.L. supervised the overall project, including the article structure, and experimental design. T.P. and C.D. contributed to the adversarial techniques and experimental design. H.G. performed all experiments and provided the draft writing materials. H.S. and C.L. extensively revised and finalized the manuscript, figures, and other materials. All authors participated in the editing of the manuscript.

\section*{Corresponding authors} Hao Sun (\url{haosun@ruc.edu.cn}) and
Chongxuan Li (\url{chongxuanli@ruc.edu.cn}).

\clearpage
\setcounter{figure}{0}
\renewcommand{\figurename}{Extended Data Figure}
\setcounter{table}{0}
\renewcommand{\tablename}{Extended Data Table}

\begin{figure*}[t!]
\begin{center}
\centerline{\includegraphics[width=0.99\linewidth]{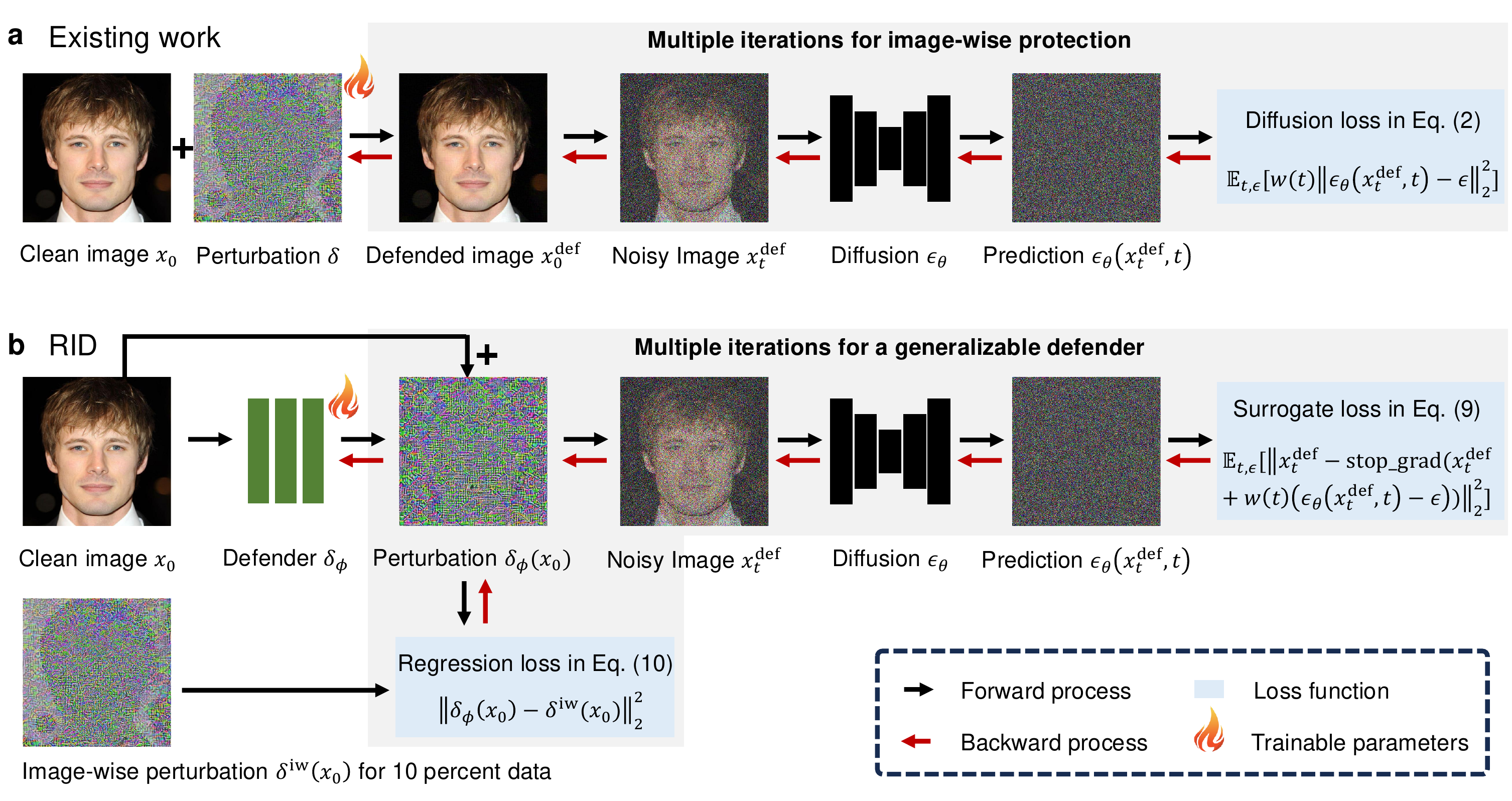}}
\vspace{-3mm}
\caption{\textbf{Comparison of training pipelines: existing optimization-based methods vs. our real-Time identity defender (RID).} \textbf{a}, Optimization-based methods individually optimize perturbations for each image via continuous gradient ascent, resulting in significant computational overhead and prolonged defense time.
\textbf{b}, RID’s training framework employs a DiT network that learns to generate image-specific permutations. This process is guided by two key loss functions: (1) adversarial score distillation loss (Adv-SDS) incorporates pre-trained model priors to increase the diffusion robustness of defended images, and (2) regression Loss matches RID-generated perturbations with those from precomputed optimization-based methods for ten percent of data. All facial images used in this figure are sourced from publicly available datasets and are permitted for academic purposes.
}
\label{fig-append2}
\end{center}
\vspace{0mm}
\end{figure*}

\clearpage
\begin{figure*}[t!]
\begin{center}
\centerline{\includegraphics[width=0.99\linewidth]{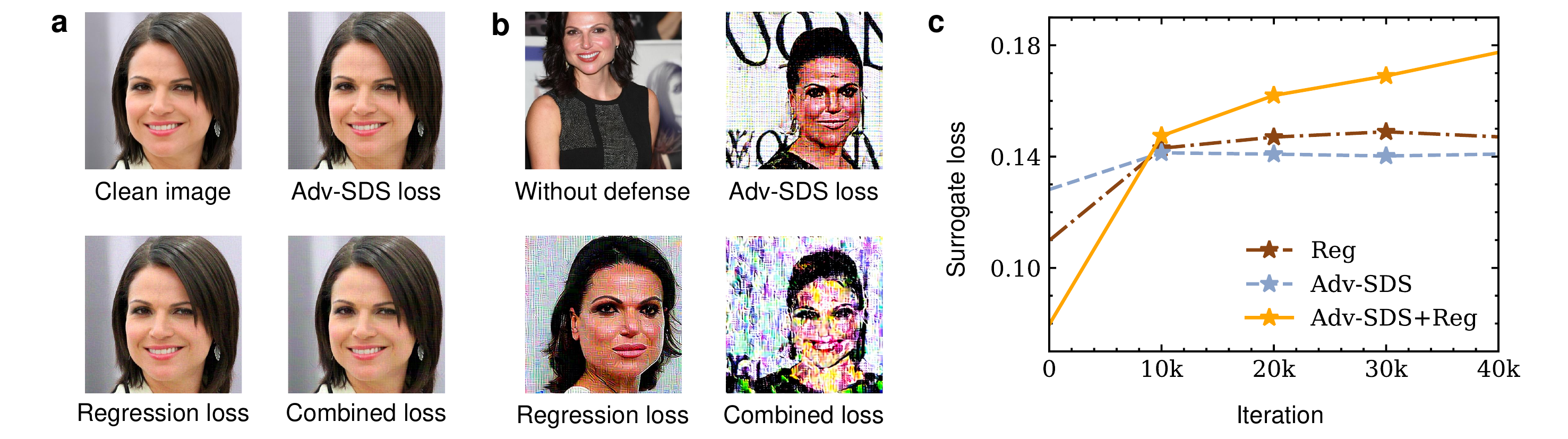}}
\caption{\textbf{The ablation study of training the RID with different losses.}
\textbf{a}, We show the defended images for all loss functions. The regularization term effectively mitigates the grid-like artifacts introduced by Adv-SDS in the protected images.
\textbf{b}, We show the generated samples from the personalization diffusion models fine-tuned on images defended by the loss functions. The combined loss achieves the best qualitative protection performance—the identity is completely obscured in the generated images.
\textbf{c}, Quantitative comparisons of diffusion losses on defended images further confirm that the combined loss consistently delivers the strongest protection. All facial images used in this figure are sourced from publicly available datasets and are permitted for academic purposes.
}
\label{fig7}
\end{center}
\vspace{0mm}
\end{figure*}

\clearpage
\begin{table}[t!]
\centering
\caption{\textbf{Model architectures and training computation for RID.}}
\label{tab:model_specs}
\small
\begin{tabular}{lcccccc}
\toprule
Model & Layers $N$ & Hidden size $d$ & Patch size & Heads & Params & Gflops\\
\midrule
RID-S (Default)  & 14 & 1152 & 16& 8  & 339.10M  & 230.29G  \\
RID-M  & 14 & 1152 & 8 & 8 & 337.77M & 915.41G \\
RID-B  & 26 & 1152 & 8 & 8 & 624.63M & 1698.12G \\
\bottomrule
\end{tabular}
\end{table}

\end{document}